\documentclass[review]{elsarticle}

\usepackage{color}
\usepackage{times}
\usepackage{epsfig}
\usepackage{graphicx}

\usepackage{amsmath}
\usepackage{amssymb}
\usepackage[ruled,vlined,linesnumbered]{algorithm2e}
\usepackage{bbm}
\usepackage{graphicx} 
\usepackage{subfigure} 
\usepackage{subfloat}
\usepackage{verbatim}
\usepackage{multirow}
\usepackage{makecell}

\usepackage{colortbl,booktabs}
\usepackage{lscape}
\usepackage{url}
\usepackage{diagbox}
\usepackage{mathtools} 
\usepackage{xcolor,colortbl}
\usepackage{dsfont}
\usepackage{picins}

\usepackage{amssymb}

\usepackage[figuresright]{rotating}

\newtheorem{definition}{Definition}
\newtheorem{proposition}{Proposition}
\newproof{pf}{Proof}

\begin{document}

\begin{frontmatter}

\title{
Self-Reflective Risk-Aware Artificial Cognitive Modeling for Robot Response to Human Behaviors
}




\author[csm]{Fei Han}
\ead{fhan@mines.edu}
\author[utk]{Christopher Reardon}
\ead{creardon@utk.edu}
\author[utk]{Lynne E. Parker}
\ead{leparker@utk.edu}
\author[csm]{Hao Zhang}
\ead{hzhang@mines.edu}

\address[csm]{Human-Centered Robotics Lab,
Department of Electrical Engineering and Computer Science,\\
Colorado School of Mines, Golden, CO 80401, USA}

\address[utk]{Distributed Intelligence Lab,
Department of Electrical Engineering and Computer Science,\\
University of Tennessee, Knoxville, TN 37996, USA}

\begin{abstract}
In order for cooperative robots (``co-robots'') to respond to human behaviors
accurately and efficiently in human-robot collaboration, interpretation of human actions, awareness of new situations, and appropriate decision making are all crucial abilities for co-robots.
For this purpose, the human behaviors should be interpreted by co-robots in the same manner as human peers. To address this issue, a novel interpretability indicator is
introduced so that robot actions are appropriate to the current human behaviors.
In addition, the complete consideration of all potential situations of a robot's environment is nearly impossible in real-world applications,
making it difficult for the co-robot to act appropriately and safely in new scenarios.
This is true even when the pretrained model is highly accurate in a known situation.
For effective and safe teaming with humans,
we introduce a new generalizability indicator that allows a co-robot to
self-reflect and reason about when an
observation falls outside the co-robot's learned model.
Based on topic modeling and two novel indicators,
we propose a new \emph{Self-reflective Risk-aware Artificial Cognitive} (SRAC) model.
The co-robots are able to consider action risks and identify new situations so that
better decisions can be made.
Experiments both using real-world datasets and on physical robots suggest that
our SRAC model significantly outperforms the traditional methodology
and enables better decision making in response to human activities.

\end{abstract}


\end{frontmatter}

\section{Introduction}


In human-robot collaboration, it is crucial for a cooperative robot (``co-robot'') to have the abilities of perception of human activities and corresponding appropriate decision-making to understand and interact with human peers. In order to provide these important capabilities, an artificial cognitive model integrating perception, reasoning, and decision making modules is required by intelligent co-robots to respond to humans effectively.
Artificial cognition has its origin in cybernetics; its intention is to create a science of mind based on logic \cite{Varela_99}.
Among other mechanisms,
cognitivism is a most widely used cognitive paradigm \cite{Vernon_TEC07}.
Several cognitive architectures were developed within this paradigm,
including ACT-R \cite{Anderson_AP96}
(and its extensions ACT-R/E \cite{trafton2013act}, ACT-R$\Phi$ \cite{dancy2013act}, etc),
Soar \cite{laird2012soar},
C4  \cite{Isla_IJCAI01},
and architectures for robotics \cite{Burghart_ICHR05}.
Because an architecture represents the connection and interaction of different cognitive components,
it cannot accomplish a specific task
on its own
without specifying each component that can provide knowledge to the cognitive architecture.
The combination of the cognitive architecture and components
is usually referred to as a {cognitive model} \cite{Vernon_TEC07}.


\begin{figure}[!ht]
\centering
\includegraphics[width= 0.8 \textwidth]{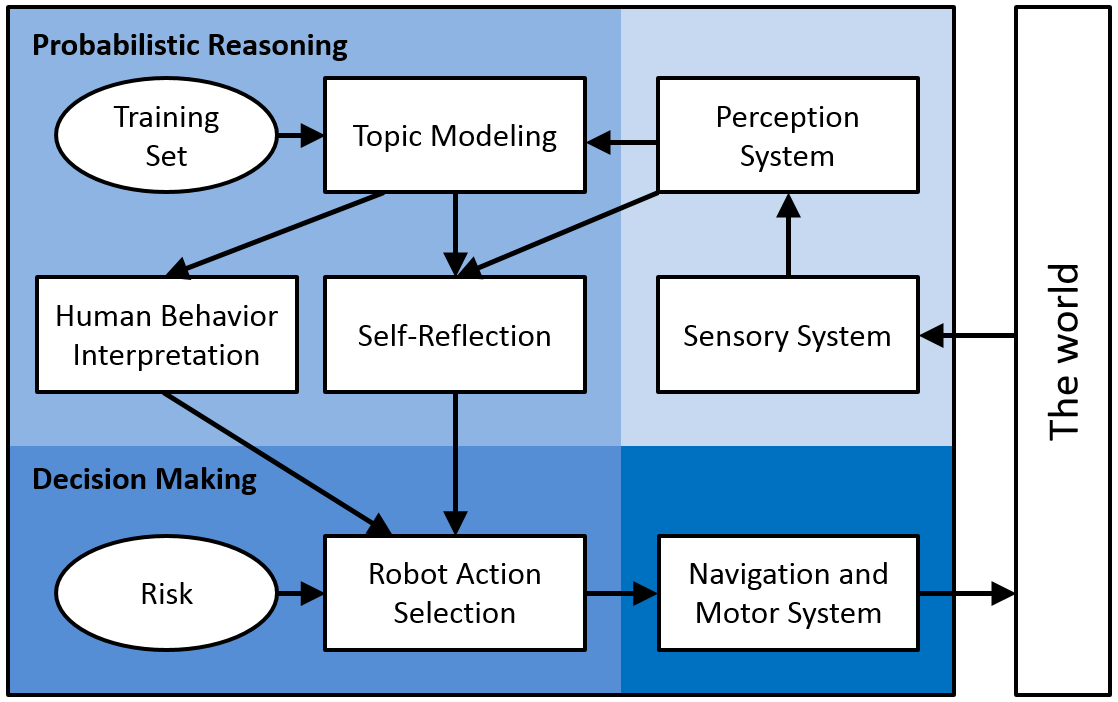}
\caption{
Overview of the SRAC model for robot response to human activities.
The novel self-reflection module allows a co-robot to reason about
when the learned knowledge no longer applies.
Decisions are made by considering both human activity category distributions
and robot action risks.
Entities in ellipses are prior knowledge to the SRAC model.
Information flows from modules with lighter colors to those with darker colors.
}\label{fig:ArchitectureCogSys}
\end{figure}


Implementing such an artificial cognitive system is challenging,
since the high-level processes (e.g., reasoning and decision making)
must be able to seamlessly work with the low-level components, e.g., perception, under significant uncertainty in a complex environment \cite{Schmid_CogSysRes12}.
In the context of human-robot collaboration,
perceiving human behaviors is a necessary component,
where uncertainty arises due to human behavior complexity,
including variations in human motions and appearances,
and challenges of machine vision,
such as lighting changes and occlusion.
This perception uncertainty is addressed in this work
using the bag-of-visual-words (BoW) representation
based on local spatio-temporal features,
which has previously shown promising performance \cite{Le_CVPR11,Alahi_CVPR12,Zhang_IROS11}.

To further process the perceptual data,
a high-level reasoning component is necessary for a co-robot to make decisions.
In recent years, topic modeling has attracted increasing attention
in human behavior discovery and recognition
due to its ability to generate a distribution over activities of interest,
and its promising performance using BoW representations in robotics applications \cite{Zhang_IROS11,Girdhar_IJRR13}.
However, previous work only aimed at human behavior understanding;
the essential task of incorporating topic modeling into cognitive decision making
(e.g., selecting a response action)
is not well analyzed.

Traditional activity recognition systems typically use accuracy as a performance metric \cite{Niebles_IJCV08}.
Because the accuracy metric ignores the distribution of activity categories,
which is richer and more informative than a single label,
it is not appropriate for decision making.
For example,
in a task of behavior understanding with two categories,
assume that two recognition systems obtain two distributions
$[0.8,0.2]$ and $[0.55, 0.45]$ on a given observation,
and the ground truth indicates the first category is correct.
Although both systems are accurate,
in the sense that the most probable category matches the ground truth,
the first model obviously performs better,
since it better separates the correct from the incorrect assignment.
Previous studies did not consider this important phenomenon.

In real-world applications, artificial cognitive models must be applied in an online fashion.
If a co-robot is unable to determine whether its knowledge is accurate, then
if it observes a new human behavior that was not presented during the training phase,
it cannot be correctly recognized,
because the learned behavior recognition model no longer applies.
Decision making based on incorrect recognition
in situations like these
can result in inappropriate or even unsafe robot action response.
Thus,
an artificial cognitive model requires the capability to self-reflect
whether the learned activity recognition system
becomes less applicable,
analogous to human self-reflection on learned knowledge,
when applied in a new unstructured environment.
This problem was not well investigated in previous works.

In this paper, we develop a novel artificial cognitive model,
based on topic models,
for robot decision making in response to human behaviors.
Our model is able to incorporate human behavior distributions
and take into account robot action risks to make more appropriate decisions (we label this ``risk-aware").
Also, our model is able to identify new scenarios when the learned recognition subsystem is less applicable (we label this ``self-reflective").
Accordingly, we call our model the \emph{self-reflective, risk-aware artificial cognitive} (SRAC) model.

Our primary contributions are twofold:
\begin{itemize}
\item
Two novel indicators are proposed.
The \emph{interpretability indicator} ($I_I$)
enables a co-robot to interpret category distributions
in a similar manner to humans.
The online \emph{generalizability indicator} ($I_G$)
measures the human behavior recognition model's generalization capacity
(i.e., how well unseen observations can be represented by the learned model).

%



\item
A novel artificial cognitive model (i.e., SRAC) is introduced
based on topic models and the indicators,
which is able to consider robot action risks and perform self-reflection
to improve robot decision making in response to human activities in new situations.

\end{itemize}

The rest of the paper is organized as follows.
We first overview the related work in Section \ref{sec:RelatedWork}.
Then, the artificial cognitive architecture
and its functional modules are described in Section \ref{sec:CognitiveSystem}.
Section \ref{sec:TopicModels} introduces the new indicators.
Section \ref{sec:DecisionMaking} presents self-reflective risk-aware decision making.
Experimental results are discussed in Section \ref{sec:ExperimentResults}.
Finally,
we conclude our paper in Section \ref{sec:Conclusion}.

\section{Related Work}\label{sec:RelatedWork}
In this section, we provide an overview of a variety of methods related to our proposed SRAC cognitive model for human-robot teaming, including human activity recognition, topic models,
and artificial cognitive modeling.

\subsection{Human Activity Recognition}

We focus our review on the commonly used sequential and space-time volume methods \cite{Aggarwal_CSUR11}.
A comprehensive review of different aspects of human activity recognition (HAR) 
is presented in \cite{Aggarwal_CSUR11} and \cite{borges2013video}.


A popular sequential method is to use centroid trajectories to identify human activities in visual data,
in which a human is represented as a single point
indicating the human's location.
Chen and Yang represented a human with just a point to derive the gait features for the pedestrian detection \cite{chen2014extraction}.
Ge \emph{et al.} extracted pedestrian trajectories from video to automatically detect small groups of people traveling together \cite{ge2012vision}.
These methods can avoid the influence of human appearances such as dresses and carrying, but are not able to recognize activities involving various relative body-part movements.
Another sequential method relies on human shapes,
including human contours and silhouettes.
Singh \emph{et al.} extracted directional vectors from the silhouette contours and utilize the distinct data distribution of these vectors in a vector space for activity recognition \cite{Singh_2008}.
Junejo \emph{et al.} transformed silhouettes of a human from every frame into time-series, then each of these time series is converted into the symbolic vector to represent actions \cite{junejo2014silhouette}.
A third method is based on body-part models.
Zhang and Parker implemented a bio-inspired predictive orientation decomposition (BIPOD) to construct representations of people from skeleton trajectories for the activity recognition and prediction, where the human body is decomposed into five body parts \cite{zhang2015bio}.
However, techniques based on shapes and/or body-part models
rely on human and body-part detection,
which are hard-to-solve problems due to occlusions and dynamic backgrounds, among others.

Space-time volume methods use local features to represent local texture and motion variations regardless of global human appearance and activity.
A large number of HAR methods are based on SIFT features \cite{Lowe_IJCV04}
and its extensions \cite{Sun_CVPR09}. 
For example, Behera \emph{et al.} proposed a random forest that unifies randomization, discriminative relationships mining and a Markov temporal structure for real-time activity recognition with SIFT features \cite{behera2014real}.
However, SIFT features only encode appearance information
and are not able to represent temporal information.
STIP features were introduced in \cite{Schuldt_ICPR04}
and SVMs were applied to classify human activities.
Dollar \emph{et al.} used separable filters in spatial and temporal dimensions to extract features for HAR \cite{Dollar_ICCV05}.
Four-dimensional features were also introduced in \cite{Zhang_IROS11} to combine depth information to classify human activities.


Previous work focused only on recognizing human activities
but did not discuss the consequent issue:
how a co-robot can make decisions based on recognition results,
especially when risks are associated with different human activities.



\subsection{Topic Models and Evaluation}


Among other machine learning techniques,
topic modeling has been widely applied to HAR.
Zhao \emph{et al.} incorporated Bayesian learning into an undirected topic model and proposed a "relevance topic model" for the unstructured social group activity recognition \cite{zhao2013relevance}.
A semi-latent topic model trained in a supervised fashion was introduced in \cite{Wang_PAMI09}
and used to classify activities in videos.
Zhang and Parker adopted topic models to classify activities in 3D point clouds from color-depth cameras on mobile robots \cite{Zhang_IROS11}.
Topic models were also widely used to discover human activities in streaming data.
The use of topic models was explored in \cite{Huynh_DAP08} to discover daily activity patterns in wearable sensor data.
An unsupervised topic model was proposed in \cite{Farrahi_TIST11} to detect daily routines from streaming location and proximity data.
Taking temporal and/or object relational information into account, Freedman \emph{et al.} explored a new method using topic models for both plan recognition and activity recognition objective \cite{freedman2014temporal}.

Although there is a significant body of work introducing and developing sophisticated topic models, 
few efforts have been undertaken to evaluate them.
Existing methods are dominated by either intrinsic methods,
(e.g., computing the probability of held-out documents to evaluate generalization ability \cite{Wallach_ICML2009}) 
or extrinsic methods that rely on external tasks,
(e.g., information retrieval \cite{Wei_ICRDIR06}). 
Some work also focused on evaluation of topic modeling's interpretability as semantically coherent concepts.
For example,
Chang \emph{et al.} demonstrated that the probability of held-out documents is
not always a good indicator of human judgment \cite{chang_nips09}.
Newman \emph{et al.} showed that metrics based on word co-occurrence statistics
are able to predict human evaluations of topic quality \cite{Newman_HLT10}.

As recently pointed out by Blei \cite{Blei_ComACM12},
topic model evaluation is an essential research topic. 
Despite this, previous works use only the accuracy metric to evaluate topic modeling results in HAR tasks;
issues such as the model's interpretability and generalizability have not been studied.
In this paper, we analyze these two aspects of topic model evaluation in HAR tasks,
explore their relationship,
and show how they can be used to improve robot decision making.

\subsection{Artificial Cognitive Modeling}

Artificial cognition has its origin in cybernetics with the intention to create a science of mind based on logic \cite{Varela_99}.
Among other cognitive paradigms, cognitivism has undoubtedly been predominant to date
\cite{Vernon_TEC07}.
Within the cognitivism paradigm,
several cognitive architectures were developed,
including
Soar \cite{laird2012soar},
ACT-R \cite{Anderson_AP96}
(and its extensions ACT-R/E \cite{trafton2013act}, ACT-R$\Phi$ \cite{dancy2013act}, etc),
C4 \cite{Isla_IJCAI01},
and architectures for robotics \cite{Burghart_ICHR05},
which are relatively independent of applications \cite{Gray_HCI97}.
Because architectures represent the mechanism for cognition but lack the relevant information for using that mechanism, they cannot accomplish anything in
their own right and need to be provided with knowledge to conduct a specific task.
The combination of a cognitive architecture and a particular knowledge set
is generally referred to as a \emph{cognitive model} \cite{Vernon_TEC07}.
The knowledge incorporated in cognitive models
is typically determined by human designers \cite{Vernon_TEC07}.
The knowledge can be also learned and adapted using machine learning techniques.

Cognitive models have been widely used in human-machine interaction and robotic vision applications.
For example, cognitive modeling was adopted in \cite{Duric_IEEE02,lallee2014efaa,baxter2013cognitive} to construct intelligent human-machine interaction systems. 
Cognitive perception systems were also used to recognize traffic signs \cite{Yang_TST13,halbrugge2013act},
interpret traffic behaviors \cite{Nagel_AIM04,lawitzky2013interactive},
and recognize human activities \cite{Crowley_CVS06,brdiczka2009learning}.
Over the last decade, probabilistic models of cognition,
as an alternative of deterministic cognitive models,
have attracted more attention in cognitive development \cite{Xu_Cog11}.
For example, an adaptive remote data mirroring system was proposed applying
dynamic decision networks in \cite{bencomo2013dynamic}.
Another cognitive model was introduced in \cite{kafai2012dynamic}
to apply dynamic Bayesian networks for vehicle classification.
Probabilistic models have also been widely used for learning and reasoning in cognitive modeling \cite{Chater_TCS06}. 

We believe we are the first to adopt topic models for the construction of reliable artificial cognitive models and show that they are particularly suited for this task.
We demonstrate topic modeling's ability to combine risks in decision making.
In addition, we develop two evaluation metrics
and show their effectiveness in model selection and decision making.
These aspects were not addressed in previous artificial cognitive modeling research.

 \section{Topic Modeling for Artificial Cognition}\label{sec:CognitiveSystem}

\subsection{Cognitive Architecture Overview}


The proposed SRAC model is inspired by
the C4 cognitive architecture \cite{Isla_IJCAI01}. 
As shown in Fig. \ref{fig:ArchitectureCogSys},
our model is organized into four modules by their functionality:
\begin{itemize}
\item \emph{Sensory and perception}:
    Visual cameras observe the environment.
    Then, the perception system
    builds a BoW representation from raw data, which can be processed by topic models.

\item \emph{Probabilistic reasoning}: 
    Topic models are applied to reason about human activities,
    which are trained off-line and used online.
    The training set is provided as a prior that encodes a history of sensory information.
    This module uses the proposed indicators to select topic models that better match human's perspective,
    and to discover new activities in an online fashion.
\item \emph{Decision making}:
    Robot action risk based on topic models and the evaluation results is estimated and
    a response robot action that minimizes this risk is selected.
    The risk is provided as a prior to the module.
\item \emph{Navigation and motor system}: 
     The selected robot action is executed in response to human activities.
\end{itemize}


\subsection{Topic Modeling}

%




Latent Dirichlet Allocation (LDA) \cite{Blei_JMLR03}, which showed promising activity recognition performance in our prior work \cite{ Zhang_IROS11}, is applied in the SRAC model.

Given a set of observations $\{\boldsymbol{w}\}$, 
LDA models each of $K$ activities as a multinomial distribution of all possible visual words
in the dictionary $\boldsymbol{D}$.
This distribution is parameterized by
$\boldsymbol{\varphi} \!=\! \{\varphi_{w_1}, \dots, \varphi_{w_{|\boldsymbol{D}|}}\}$,
where $\varphi_{w}$ is the probability that the word ${w}$ is generated by the activity.
LDA also represents each $\boldsymbol{w}$ as a collection of visual words,
and assumes that each word $w \in \boldsymbol{w}$ is associated with a latent activity assignment $z$.
By applying the visual words to connect observations and activities,
LDA models $\boldsymbol{w}$ as a multinomial distribution over the activities,
which is parameterized by $\boldsymbol{\theta} \!=\! \{\theta_{z_1},\dots,\theta_{z_K}\}$,
where $\theta_{z}$ is the probability that $\boldsymbol{w}$ is generated by the activity $z$.
LDA is a Bayesian model,
which places Dirichlet priors on the multinomial parameters:
$\boldsymbol{\varphi} \!\sim\! \operatorname{Dir}(\boldsymbol{\beta})$ and
$\boldsymbol{\theta} \!\sim\! \operatorname{Dir}(\boldsymbol{\alpha})$,
where $\boldsymbol{\beta} \!=\! \{\beta_{w_1},\dots,\beta_{w_{|\boldsymbol{D}|}}\}$
and $\boldsymbol{\alpha} \!=\! \{\alpha_{z_1},\dots,\alpha_{z_K}\}$ are
the concentration hyperparameters.

%

One of the major objectives in HAR tasks to is to estimate the parameter $\boldsymbol{\theta}$,
i.e., the per-observation activity proportion.
However, exact parameter estimation is intractable in general \cite{Blei_JMLR03}.
Our model applies Gibbs sampling \cite{Griffiths_NAS04} to compute the \emph{per-observation activity distribution} $\boldsymbol{\theta}$,
based on two considerations:
1) This sampling-based method is generally accurate,
since it asymptotically approaches the correct distribution \cite{Porteous_KDD08}, and
2) This method can be used to intrinsically evaluate topic model's performance \cite{Wallach_ICML2009},
thereby providing a consistent method to infer, learn, and evaluate topic models.
At convergence,
the element $\theta_{z_k} \!\in\! \boldsymbol{\theta}$, $k \!=\! 1,\dots,K$, is estimated by:
\begin{eqnarray}
\hat{\theta}_{z_k} = \frac{n_{z_k} + \alpha_{z_k}}{\sum_{z}{(n_z + \alpha_z)}},
\end{eqnarray}
where $n_z$ is the number of times that a visual word is assigned to activity $z_k$ in the observation.



The incorporation of topic models into cognitive modeling has several important advantages.
First, as a probabilistic reasoning approach,
it serves as a bridge to allow information to flow from the perception module to the decision making module.
Second, the ability to model per-observation activity distribution allows topic models
to take into account the risks of all robot actions in a probabilistic way
and make an appropriate decision.
Third, by introducing an extrinsic evaluation metric for topic model selection,
the constructed cognitive system is able to accurately interpret human activities.
Fourth, the unsupervised nature of topic modeling, which is explored using our new intrinsic metric,
facilitates online discovery of new knowledge (e.g., human activities).
All these advantages allow us to apply topic models to construct an artificial cognitive system
that is able to better interpret human activities,
discover new knowledge and react more appropriately and safely to humans,
which is highly desirable for real-world online human-robot interaction scenarios.

\section{Interpretability and Generalizability}\label{sec:TopicModels}

To improve artificial cognitive modeling,
we introduce two novel indicators and discuss their relationship in this section,
which are the core of the \emph{Self-Reflection} module in Fig. \ref{fig:ArchitectureCogSys}.

\subsection{Interpretability Indicator} \label{subsec:ii}

We observe that accuracy is not an appropriate assessment metric for robot decision making,
since it only considers the most probable human activity category and ignores the others.
To utilize the category distribution, which contains much richer information,
the \emph{interpretability indicator}, denoted by $I_I$, is introduced.
$I_I$ is able to encode how well topic modeling matches human common sense.
Like the accuracy metric, $I_I$ is an extrinsic metric,
meaning that it requires a ground truth to compute.
Formally, $I_I$ is defined as follows:

\begin{definition}[Interpretability indicator]\label{def:II}
Given the observation $\boldsymbol{w}$
with the ground truth $g$ and the distribution $\boldsymbol{\theta}$ over $K \geq 2$ categories,
let $\boldsymbol{\theta}_s = (\theta_1,$ $\dots, \theta_{k-1}, \theta_k,$ $\theta_{k+1}, \dots, \theta_K)$ denote
the sorted proportion satisfying
$\theta_1 \geq \cdots \geq \theta_{k-1} \geq \theta_k \geq \theta_{k+1} \geq \cdots \geq \theta_K \geq 0$
and $\sum_{i=1}^{K} \theta_i = 1$,
and let $k \in \{1,\cdots, K\}$ represent the index of the assignment in $\boldsymbol{\theta}_s$ that matches $g$.
The interpretability indicator $I_I (\boldsymbol{\theta},g) = I_I (\boldsymbol{\theta}_s,\!k)$ is defined as:
\begin{eqnarray} \label{eq:II}
\small{
I_I (\boldsymbol{\theta}_s,\!k) \triangleq
\frac{1}{a}
\left( \frac{K \!-\! k}{K \!-\! 1} + \mathds{1}(k \!=\! K) \right)
\left( \frac{\theta_k}{\theta_1} \!-\!
\frac{\theta_{k + \mathds{1}(k \neq  K)}}{\theta_k} \!+\! b  \right)
\!\!
}
\end{eqnarray}
where $\mathds{1}(\cdot)$ is the indicator function, and $a=2$, $b=1$ are normalizing constants.
\end{definition}

The indicator $I_I$ is defined over the per-observation category proportion $\boldsymbol{\theta}$,
which takes values in the $(K\!-\!1)$-simplex \cite{Blei_JMLR03}.
The sorted proportion $\boldsymbol{\theta}_s$ is computed through sorting $\boldsymbol{\theta}$,
which is inferred by topic models.
In the definition, the ground truth is represented by its location in $\boldsymbol{\theta}_s$,
i.e., the $k$-th most probable assignment in $\boldsymbol{\theta}_s$ matches the ground truth label.
The indicator function $\mathds{1}(\cdot)$ in Eq. (\ref{eq:II}) is adopted to deal with the special case when $k = K$.

For an observation in an activity recognition task with $K$ categories,
given its ground truth index $k$ and sorted category proportion $\boldsymbol{\theta}_s$,
we summarize $I_I$'s properties as follows:

\begin{proposition}[$I_I$'s properties]\label{prop1}

The interpretability indicator $I_I(\boldsymbol{\theta},g) = I_I(\boldsymbol{\theta}_s,k)$ satisfies the following properties:

1. If $k = 1$, $\forall \boldsymbol{\theta}_s$, $I_I(\boldsymbol{\theta}_s,k) \geq 0.5$.

2. If $k = K$, $\forall \boldsymbol{\theta}_s$, $I_I(\boldsymbol{\theta}_s,k) \leq 0.5$.

3. $\forall \boldsymbol{\theta}_s$, $I_I(\boldsymbol{\theta}_s,k) \in [0,1]$.

4. $\forall k \in \{1,\dots, K\}$ and $\boldsymbol{\theta}_s$, $\boldsymbol{\theta}'_s$ such that $\theta_1 \geq\theta_1'$, $\theta_k = \theta_k'$ and $\theta_{k+\mathds{1}(k \neq K)} = \theta_{k+\mathds{1}(k \neq K)}'$,
$I_I (\boldsymbol{\theta}_s,k) \leq I_I (\boldsymbol{\theta}'_s,k)$ holds.

5. $\forall k \in \{1,\dots, K\}$ and $\boldsymbol{\theta}_s$, $\boldsymbol{\theta}'_s$ such that $\theta_{k+\mathds{1}(k \neq K)} \geq\! \theta_{k + \mathds{1}(k \neq K)}'$,
$\theta_1 \!=\! \theta_1'$ and
$\theta_k \!=\! \theta_k'$,
$I_I (\boldsymbol{\theta}_s,k) \leq I_I (\boldsymbol{\theta}'_s,k)$ holds.

6. $\forall k \in \{1,\dots, K\}$ and $\boldsymbol{\theta}_s$, $\boldsymbol{\theta}'_s$ such that
$\theta_k \geq \theta_k'$,
$\theta_1 = \theta_1'$ and
$\theta_{k+\mathds{1}(k \neq K)} = \theta_{k + \mathds{1}(k \neq K)}'$,
$I_I (\boldsymbol{\theta}_s,k) \geq I_I (\boldsymbol{\theta}'_s,k)$ holds.

7. $\forall k, k' \in \{1,\dots, K\}$ such that $k \leq k' < K$ and
$\forall \boldsymbol{\theta}_s$, $\boldsymbol{\theta}'_s$ such that
$\theta_k = \theta_k'$,
$\theta_1 = \theta_1'$ and
$\theta_{k+\mathds{1}(k \neq K)} = \theta_{k + \mathds{1}(k \neq K)}'$,
$I_I (\boldsymbol{\theta}_s,k) \geq I_I (\boldsymbol{\theta}'_s,k')$ holds.

\end{proposition}

\begin{pf}
See \ref{appendix:1}.
\end{pf}

The indicator $I_I$ is able to quantitatively measure
how well topic modeling can match human common sense,
because it captures three essential considerations to
simulate the process of how humans evaluate the category proportion $\boldsymbol{\theta}$:
\begin{itemize}
\item
A topic model performs better, in general, if it obtains a larger $\theta_k$ (Property 6).
In addition, a larger $\theta_k$ generally indicates
$\theta_k$ is closer to the beginning in $\boldsymbol{\theta}_s$
and further away from the end (Property 7). 

\textit{Example}:
A topic model obtaining the sorted proportion
$[0.4, \boxed{0.35}, 0.15, 0.10]$
performs better than a model obtaining $[0.4, \boxed{0.30}, 0.15, 0.15]$,
where the ground truth is marked with a box, i.e., $k = 2$ in the example.

\item A smaller difference between $\theta_k$ and $\theta_1$ indicates better modeling performance (Properties 4 and 5), in general.
Since the resulting category proportion is sorted,
a small difference between $\theta_k$ and $\theta_1$ guarantees $\theta_k$ has an even smaller difference from $\theta_2$ to $\theta_{k-1}$.

\textit{Example}: A topic model obtaining the sorted proportion
$[0.4, \boxed{0.3}, 0.2, 0.1]$
performs better than the model with the proportion $[0.5, \boxed{0.3}, 0.2, 0]$.

\item A larger distinction between $\theta_k$ and $\theta_{k+1}$ generally indicates better modeling performance (Properties 5 and 6),
since it better separates the correct assignment from the incorrect assignments with lower probabilities.

\textit{Example}: A topic model obtaining the sorted proportion
$[0.4, \boxed{0.4}, 0.1, 0.1]$ performs better than the topic model obtaining the proportion
$[0.4, \boxed{0.4}, 0.2, 0]$.

\end{itemize}

The indicator $I_I$ extends the \emph{accuracy} metric $I_A$
(i.e., rate of correctly recognized data),
as described in Proposition \ref{prop2}:

\begin{proposition}[Relationship of $I_I$ and $I_A$] \label{prop2}
The accuracy measure $I_A$ is a special case of $I_I(\boldsymbol{\theta}_s, k)$, when
$\theta_1 = 1.0$, $\theta_2 =\! \dots \!=  \theta_K = 0$,
and $k = 1$ or $k = K$.
\end{proposition}

\begin{pf}
See \ref{appendix:2}.
\end{pf}



\subsection{Generalizability Indicator} \label{subsec:ig}

An artificial cognitive model requires the crucial capability of detecting new situations and being aware that the learned knowledge becomes less applicable
in an online fashion. 
To this end, we propose the \emph{generalizability indicator} ($I_G$),
an intrinsic metric
that does not require ground truth to compute and consequently can be used online.

The introduction of $I_G$ is inspired by the perplexity metric
(also referred to as held-out likelihood),
which evaluates a topic model's generalization ability
on a fraction of held-out instances using cross-validation \cite{Musat_IJCAI11}
or unseen observations \cite{Blei_NIPS05}.
The perplexity is defined as the log-likelihood of words in an observation \cite{Wallach_ICML2009}.
Because different observations may contain a different number of visual words,
we compute the \emph{Per-Visual-Word Perplexity ($Pvwp$)}.
Mathematically, given the trained topic model $\mathcal {M}$ and an observation $\boldsymbol{w}$, $Pvwp$ is defined as follows:
\begin{eqnarray}\label{eq:pvwp}
\small{
Pvwp(\boldsymbol{w} | \mathcal{M}) \!=\! \frac{1}{N} \log\! P(\boldsymbol{w} | \mathcal{M})
\!=\! \frac{1}{N} \log\! \prod_{n = 1}^{N} \!P(w_n| \boldsymbol{w}_{<n}, \mathcal{M})\!\!\!\!
}
\end{eqnarray}
where $N = |\boldsymbol{w}|$ is the number of visual words in $\boldsymbol{w}$,
and the subscript $<\!n$ denotes positions before $n$.
Because $P(\boldsymbol{w} | \mathcal{M})$ is a probability that satisfies $P(\boldsymbol{w} | \mathcal{M}) \!\leq\! 1$,
it is guaranteed $Pvwp(\boldsymbol{w}|\mathcal{M}) \!\leq\!0$.
The left-to-right algorithm,
presented in Algorithm \ref{alg:pvwpEstimation}, is used to estimate $Pvwp$,
which is an accurate and efficient Gibbs sampling method to estimate perplexity \cite{Wallach_ICML2009}.
The algorithm decomposes $P(\boldsymbol{w} | \mathcal{M})$ in an incremental, left-to-right fashion,
where the subscript $\neg n$ is a quantity that excludes data from the $n$th position.
Given  observations $\mathcal{W} \!=\! \{\boldsymbol{w}_1,\dots,\boldsymbol{w}_M\}$,
$Pvwp(\mathcal{W}| \mathcal{M})$ is defined as the average of each observation's perplexity:
\begin{eqnarray}
Pvwp(\mathcal{W} | \mathcal{M}) = \frac{1}{M} \sum_{m = 1}^{M} Pvwp(\boldsymbol{w}_m | \mathcal{M})
\end{eqnarray}

\begin{algorithm}
\SetAlgoLined
\SetKwData{Left}{left}\SetKwData{This}{this}\SetKwData{Up}{up}
\SetKwFunction{Union}{Union}\SetKwFunction{FindCompress}{FindCompress}
\SetKwInOut{Input}{Input}\SetKwInOut{Output}{Output}
\SetNlSty{textrm}{}{:}
\SetKwComment{tcc}{/*}{*/} 


\small{

\Input{
        $\boldsymbol{w}$ (observation), $\mathcal{M}$ (trained topic model), and
        $R$ (number of particles)
       }
\Output{
        $Pvwp(\boldsymbol{w}|\mathcal{M})$
}
\BlankLine

Initialize $l = 0$ and $N = |\boldsymbol{w}|$;

\For{each position $n = 1$ to $N$ in $\boldsymbol{w}$}{

Initialize $p_n = 0$;

\For{each particle $ r = 1$ to $R$}{

\For{$n' < n$}{

Sample $z_{n'}^{(r)} \sim P(z_{n'}^{(r)} | w_{n'},  \{ \boldsymbol{z}_{<n}^{(r)} \}_{\neg n'}, \mathcal{M})$;

} 

Compute $p_n = p_n + \sum_t P(w_n, z_n^{(r)} = t | z_{<n}^{(r)}, \mathcal{M})$;

Sample $z_n^{(r)} \sim P(z_n^{(r)} | w_n, z_{<n}^{(r)}, \mathcal{M})$;

} 

Update $p_n = \frac{p_n}{R}$ and $l = l + \log p_n$;

} 


\Return{ $Pvwp(\boldsymbol{w}|\mathcal{M}) \simeq \frac{l}{N}$. }

} 

\caption{Left-to-right $Pvwp$ estimation}\label{alg:pvwpEstimation}

\end{algorithm}

Based on $Pvwp$,
the generalizability indicator $I_G$,
on previously unseen observations in the testing phase or
using the held-out instances in cross-validation,
is defined as follows:
\begin{definition}[Generalizability indicator]
Let $\mathcal{M}$ denote a trained topic model,
$\mathcal{W}_{\text{valid}}$ denote the validation dataset that is used in the training phase,
and $\boldsymbol{w}$ be an previously unseen observation.
We define the generalizability indicator:
\begin{eqnarray}\label{eq:IG}
\footnotesize{
I_G(\boldsymbol{w}) \!\triangleq\!
\begin{dcases}
\frac{\exp(Pvwp(\boldsymbol{w}| \mathcal{M}))}{c \cdot \exp(Pvwp(\mathcal{W}_{valid}| \mathcal{M}))}  \\
     \quad\;\;
     {\text{if }
     \footnotesize{ \exp(Pvwp(\boldsymbol{w}|\mathcal{M})) \!<\! c \!\cdot\! \exp(Pvwp(\mathcal{W}_{valid}| \mathcal{M})) }
     } \\
  \; \text{$1$} \quad
    \!\!{\text{if } \footnotesize{ \exp(Pvwp(\boldsymbol{w}|\mathcal{M})) \!\geq\! c\! \cdot\! \exp(Pvwp(\mathcal{W}_{valid}|\mathcal{M}))}}
\end{dcases}\!\!\!\!
}
\end{eqnarray}
where $c \in [1, \infty)$ is a constant encoding novelty levels.
\end{definition}

Besides considering the topic model's generalization ability, 
$I_G$ also evaluates whether previously unseen observations are well-represented by the training set,
i.e., whether the training set used to train the topic model is exhaustive. 
The training set is defined as \emph{exhaustive} when it contains instances from all categories
that can possibly be observed in the testing phase \cite{Dundar_ICML12}.
When some categories are missing and not represented by the training set,
it is defined as \emph{non-exhaustive}; in this case, novel categories emerge in the testing phase.
Since it is impractical, often impossible, to define an exhaustive training set,
mainly because some of the categories may not exist at the time of training,
the ability to discover novelty and be aware that the learned model
is less applicable is essential for safe, adaptive decision making.
The indicator $I_G$ provides this ability through evaluating
how well new observations are represented by the validation set in the training phase.

We constrain $I_G$'s value in the range $(0, 1]$, with a greater value indicating less novelty,
which means an observation can be better encoded by the training set and the topic model generalizes better on this observation.
The constant $c$ in Eq. (\ref{eq:IG}) provides the flexibility to encode the degree to which we
consider an observation to be novel.

\subsection{Indicator Relationship} 

While the interpretability indicator interprets human activity distributions
in a way that is similar to human reasoning,
the generalizability indicator endows a co-robot with the self-reflection capability.
We summarize their relationship
in the cases when a training set is exhaustive
(i.e., training contains all possible categories)
and non-exhaustive
(i.e., new human behavior occurs during testing), as follows:

\emph{Observation (Relationship of $I_G$ and $I_I$)}:
Let $\mathcal{W}_{train}$ be
the training dataset used to train a topic model,
and $I_I$ and $I_G$ be the model's interpretability and generalizability indicators. 
\begin{itemize}
\item If $\mathcal{W}_{train}$ is exhaustive,
then $I_G \rightarrow 1$ and $I_I$ is generally independent of $I_G$. 
\item If $\mathcal{W}_{train}$ is non-exhaustive,
then $I_G$ takes values that are much smaller than $1$; $I_I$ also takes small values and is moderately to strongly correlated with $I_G$.
\end{itemize}

\begin{table}[tbh]
\caption{
Meaning and relationship of $I_I$ and $I_G$.
The gray area denotes that the situation is generally impossible.
} \label{tab:ImplicationIiIg}
\begin{center}
\vspace{-8pt}
\begin{tabular}{ r|p{3.5cm}|p{4.5cm}| }
\multicolumn{1}{r}{}
 &  \multicolumn{1}{c}{$I_G$: low}
 & \multicolumn{1}{c}{$I_G$: high} \\
\cline{2-3}
$I_I$:$\;$ low & Category is novel \newline Model is \emph{not} applicable  & Category is \emph{not} novel \newline Model is \emph{not} well interpreted \\
\cline{2-3}
$I_I$: high & \cellcolor{black!25} & Category is \emph{not} novel \newline Model is well interpreted \\
\cline{2-3}
\end{tabular}
\end{center}
\end{table}

This observation answers the critical question of
whether a better generalized topic model can lead to better recognition performance.
Intuitively, if $\mathcal{W}_{train}$ is non-exhaustive and a previously unseen observation $\boldsymbol{w}$ belongs to a novel category,
which is indicated by a small $I_G$ value,
a topic model trained on $\mathcal{W}_{train}$ cannot accurately classify $\boldsymbol{w}$.
On the other hand, if $\boldsymbol{w}$ belongs to a category that is known in $\mathcal{W}_{train}$,
then $I_G \!\rightarrow\! 1$ and the recognition performance over $\boldsymbol{w}$ only depends on
the model's performance on the validation set used in the training phase.
The meaning and relationship of the indicators $I_I$ and $I_G$ are summarized in Table \ref{tab:ImplicationIiIg},
where the gray area denotes that it is
generally impossible for a topic model
to obtain a low generalizability but a high interpretability,
as a model is never correct when presented with a novel activity.





\begin{table}[tb]
\caption{Risk levels as prior knowledge to our cognitive model.}\label{tab:RiskLevel}
\vspace{-8pt}
{
\begin{center}
\tabcolsep=0.2cm
\begin{tabular}{|l|c|l|}
\hline
 \quad Levels & Values & $\qquad\qquad\qquad$ Definition\\
 \hline\hline
Low risk    &  [1,30]	& Unsatisfied with the robot's performance.\\
\hline
Medium risk    &  [31,60]	& Annoyed or upset by the robot's actions.\\
\hline
High risk    &  [61,90]	& Interfered with, interrupted, or obstructed.\\
\hline
Critical risk    &  [95,100]	& Injured or worse (i.e., a safety risk).\\
\hline
\end{tabular}
\end{center}
}
\end{table}

\section{Self-Reflective Risk-Aware Decision Making} \label{sec:DecisionMaking}

Another contribution of this research is a decision making framework
that is capable of incorporating activity category distribution,
robot self-reflection (enabled by the indicators),
and co-robot action risk,
which is realized in the module of \emph{Decision Making}
in Fig. \ref{fig:ArchitectureCogSys}.
Our new self-reflective risk-aware decision making algorithm is presented in Algorithm \ref{alg:Decision}.

\begin{algorithm}
\SetAlgoLined
\SetKwData{Left}{left}\SetKwData{Name}{Name}\SetKwData{Up}{up}
\SetKwFunction{KwFn}{Fn}
\SetKwInOut{Input}{Input}
\SetKwInOut{Output}{Output}
\SetKwInOut{Func}{Func.}
\SetNlSty{textrm}{}{:}
\SetKwComment{tcc}{/*}{*/} 
\SetKwFunction{Union}{Union}

{

\Input{
        $\boldsymbol{w}$ (observation), $\mathcal{M}$ (trained topic model), and
        $\mathcal{N}$ (decision making bipartite network)
       }
\Output{
        $a^{\star}$ (Selected robot action with minimum risk)
}
\BlankLine

Estimate per-observation activity proportion $\boldsymbol{\theta}$ of $\boldsymbol{w}$;

Compute generalizability indicator $I_G(\boldsymbol{w})$;

\For{each robot action $ i = 1$ to $S$}{

Estimate activity-independent risk:
$r^{in}_i \!=\! \frac{1}{K}\sum_{j = 1}^{K} \! r_{ij}$;

Calculate activity-dependent risk:
$r^{de}_i = \sum_{j = 1}^{K} (\theta_j \cdot r_{ij})$;

Combine activity-independent and dependent risks,
and assign to per-observation action risk vector:
$\boldsymbol{r}^a(i) = (1 - I_G(\boldsymbol{w})) \cdot r_i^{in} + I_G(\boldsymbol{w}) \cdot r_i^{de}$;

} 

Select optimal robot action $a^{\star}$ with minimum risk in $\boldsymbol{r}^a$;

\Return{ $a^{\star}$. }

} 

\caption{Self-reflective risk-aware decision making}\label{alg:Decision}

\end{algorithm}


Given the robot action set $\boldsymbol{a} = \{a_1, \dots, a_S \}$
and the human activity set $\boldsymbol{z} = \{z_1, \dots, z_K\}$,
an action-activity risk $r_{ij}$ is defined as the amount of discomfort, interference, or harm that can be expected to occur
during the time period if the robot takes a specific action $a_i, \forall i \in \{1,\dots,S\}$
in response to an observed human activity $z_j, \forall j \in \{1,\dots,K\}$.
While $\boldsymbol{\theta}$ and $I_G$ are computed online,
the risks $\boldsymbol{r} = \{r_{ij}\}_{S \times K}$, with each element $r_{ij} \in [0, 100]$,  are manually estimated off-line by domain experts
and used as a prior in the decision making module.
In practice, the amount of risk is categorized into a small number of risk levels for simplicity's sake.
To assign a value to $r_{ij}$, a risk level is first selected.
Then, a risk value is determined within that risk level.
As listed in Table \ref{tab:RiskLevel},
we define four risk levels with different risk value ranges in our application.
We intentionally leave a five-point gap between critical risk and high risk
to increase the separation of critical risk from high risk actions.


\begin{figure}[h]
\centering
\includegraphics[width= 0.6 \textwidth]{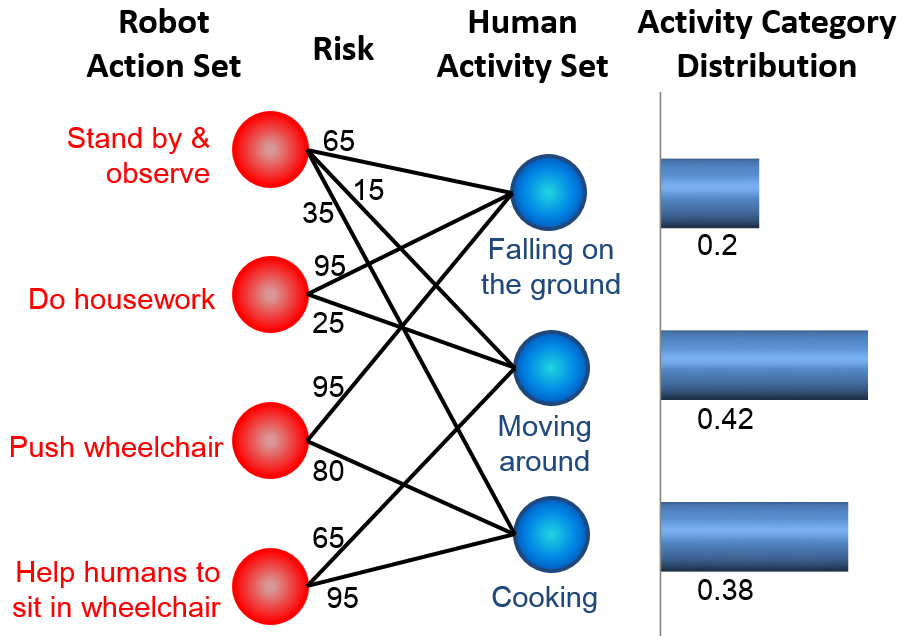}
\caption{
An illustrative example of a bipartite network (left) and the per-observation activity distribution (right) in assistive robotics applications.
}\label{fig:RiskExample}
\end{figure}

A bipartite network
$\mathcal{N} \!=\! \{\boldsymbol{a}, \boldsymbol{z}, \boldsymbol{r}\}$
is proposed to graphically illustrate the risk matrix $\boldsymbol{r}$
of robot actions $\boldsymbol{a}$ associated with human activities $\boldsymbol{z}$.
In this network, vertices are divided into two disjoint sets $\boldsymbol{a}$ and $\boldsymbol{z}$,
such that every edge with a weight $r_{ij}$ connects a vertex $a_i \in \boldsymbol{a}$ to a vertex $z_j \in \boldsymbol{z}$.
An example of such a bipartite network is depicted in Fig. \ref{fig:RiskExample} for assistive robotics applications.
The bipartite network also has a tabular representation (for example, in Table \ref{tab:RiskMat}).
Given the bipartite network,
for a new observation $\boldsymbol{w}$,
after $\boldsymbol{\theta}$ 
and $I_G(\boldsymbol{w})$ are computed in the probabilistic reasoning module,
the robot action $a^\star \in \boldsymbol{a}$ is selected according to:
\begin{eqnarray}\label{eq:ActionSelection}
\small{
a^\star \!=\! \mathop{\arg \min}\limits_{a_i:i = 1, \dots, S}
\left(
\frac{1 \!-\! I_G(\boldsymbol{w})}{K} \!\cdot\! \sum_{j=1}^{K} r_{ij} +
I_G(\boldsymbol{w}) \!\cdot\! \sum_{j = 1}^{K}{(\theta_j \!\cdot\! r_{ij})}
\right) \!\!
}
\end{eqnarray}

The risk of taking a specific robot action is determined by two separate components:
activity-independent and activity-dependent action risks.
The activity-independent risk (that is $\frac{1}{K}\sum_{j=1}^{K} r_{ij}$) measures the inherent risk of an action,
which is independent of the human activity context information,
i.e., computing this risk does not require the category distribution.
For example, the robot action ``standing-by'' generally has a smaller risk than ``moving backward'' in most situations.
The activity-dependent risk (that is $\sum_{j = 1}^{K}{(\theta_j \!\cdot\! r_{ij})}$) is the average risk weighted by context-specific information (i.e., the human activity distribution).
The combination of these two risks is controlled by $I_G$,
which automatically encodes preference over robot actions.
When the learned model generalizes well over $\boldsymbol{w}$, i.e., $I_G(\boldsymbol{w}) \!\rightarrow\! 1$,
the decision making process prefers co-robot actions that are more appropriate to the recognized human activity.
Otherwise, if the model generalizes poorly,
indicating new human activities occur
and the learned model is less applicable,
our decision making module
would ignore the recognition results
and select co-robot actions with lower activity-independent risk.


%
%
%
%

\section{Experiments} \label{sec:ExperimentResults}


\subsection{Datasets and Visual Features}

\begin{figure}[h]
\centering
  \subfigure[Weizmann dataset]{
    \label{fig:Weizmann}
    \begin{minipage}[b]{1\textwidth}
      \centering
        \includegraphics[width=1\textwidth]{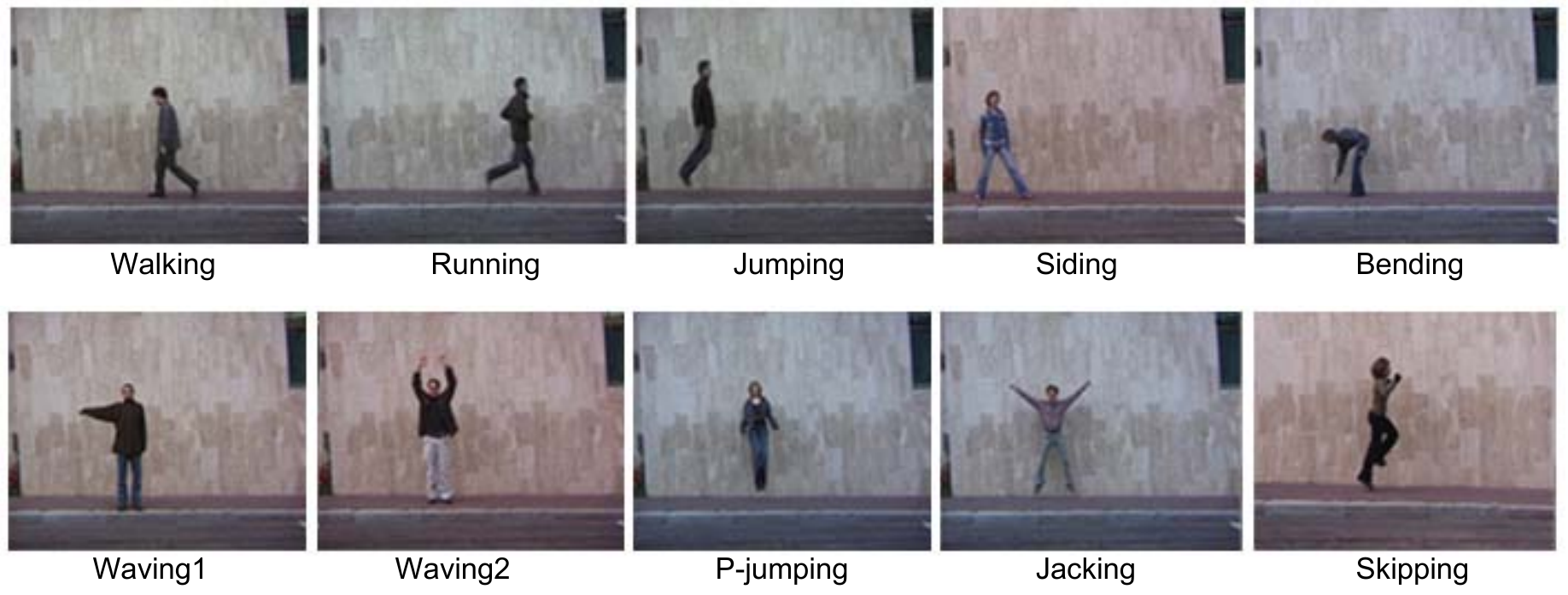}
    \end{minipage}}
  \subfigure[KTH dataset]{
    \label{fig:KTH}
    \begin{minipage}[b]{1\textwidth}
      \centering
        \includegraphics[width=1\textwidth]{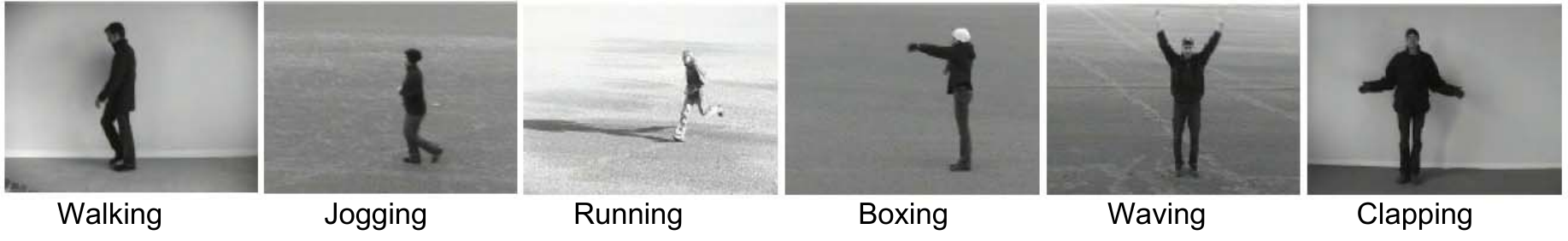}
    \end{minipage}}
      \subfigure[UTK3D dataset (3D view)]{
    \label{fig:UTK}
    \begin{minipage}[b]{1\textwidth}
      \centering
        \includegraphics[width=1\textwidth]{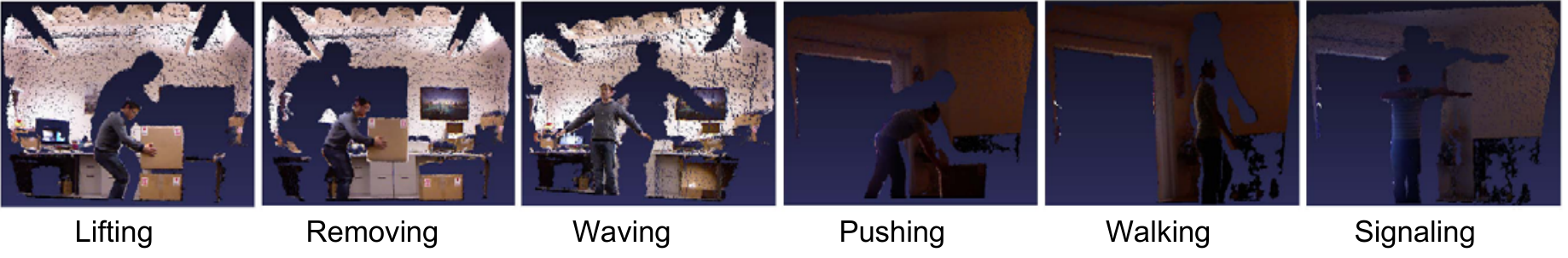}
    \end{minipage}}
  \caption{
Exemplary frames of actions in the datasets used in our experiments.
}
\label{fig:ActionDataset}
\end{figure}

We employ three real-world benchmark datasets to evaluate our cognitive model on HAR tasks,
which are widely used in the machine vision community:
the Weizmann activity dataset \cite{Gorelick_PAMI07}, the KTH activity dataset  \cite{Laptev_IJCV05},
and the UTK 3D activity dataset \cite{Zhang_IROS11}.
Illustrative examples from each activity category in the datasets are depicted in Fig. \ref{fig:ActionDataset}.

In our experiments,
we apply different types of local visual features to encode these datasets.
For 2D datasets that contain only color videos
(i.e., the Weizmann and KTH datasets),
we use two different features:
scale-invariant feature transform (SIFT) features \cite{Lowe_IJCV04}
and
space-time interest points (STIP) features \cite{Laptev_IJCV05}.
For 3D datasets that contain both color and depth videos (i.e., the UTK dataset),
we use the 4-dimensional local spatio-temporal features (4D-LSTF) \cite{Zhang_IROS11}.

SIFT features are the most commonly applied local visual features
and have desirable characteristics including invariance to transformation,
rotation and scale, and robustness to partial occlusion \cite{Lowe_IJCV04}.
We employ the algorithm and implementation in \cite{Lowe_IJCV04} to detect and describe SIFT features.
A disadvantage of SIFT features in HAR tasks is that these features are extracted in a frame-by-frame fashion,
i.e., SIFT features do not capture any temporal information.
To encode time information,
we also apply STIP along with the histogram of oriented gradients (HOG)
and histogram of optical flow (HOF) descriptors \cite{Laptev_IJCV05}. 
These two types of features are extracted using only color or intensity information.
Previous work has demonstrated that local features incorporating both depth and color information can greatly improve recognition accuracy \cite{Zhang_IROS11}.
Therefore, for the 3D UTK dataset we use 4D-LSTF \cite{Zhang_IROS11} features,
which are highly robust and distinct and are generated using both color and depth videos.
It is also noteworthy that SIFT and STIP features can be directly extracted from color or depth videos in the 3D dataset.

These feature extraction algorithms generate a collection of feature vectors for each visual observation.
Then, the feature vectors are clustered into discrete visual words using the $k$-means algorithm, 
and the number of clusters is set equal to the dictionary size.
Lastly, each feature vector is indexed by a discrete word that represents cluster assignment.
At this point,
each observation is encoded by a BoW representation, which can be perceived by topic modeling.
Although we only test the most widely used features,
one should note that
our artificial cognitive model is capable of incorporating different types of local visual features,
since our reasoning and decision making process is independent of the features given their BoW representation.


\subsection{Activity Recognition} \label{sec:experiment-II}

We first evaluate the SRAC model's capability to recognize human activities using the interpretability indicator $I_I$, when the training set is exhaustive.
In this experiment, each dataset is split into disjoint training and testing sets.
We randomly select 75\% of data instances in each category as the training set,
and employ the rest of the instances for testing.
During training,
fourfold cross-validation is used to estimate model parameters.
Then, the interpretability of the topic model is computed using the testing set,
which is fully represented by the training set and does not contain novel human activities.
This training-testing process is repeated five times to obtain reliable results.

\begin{figure}[!htb]
  \subfigure[Weizmann dataset]{
    \label{fig:ii_weizmann}
    \begin{minipage}[b]{0.5\textwidth}
      \centering
        \includegraphics[width=1\textwidth]{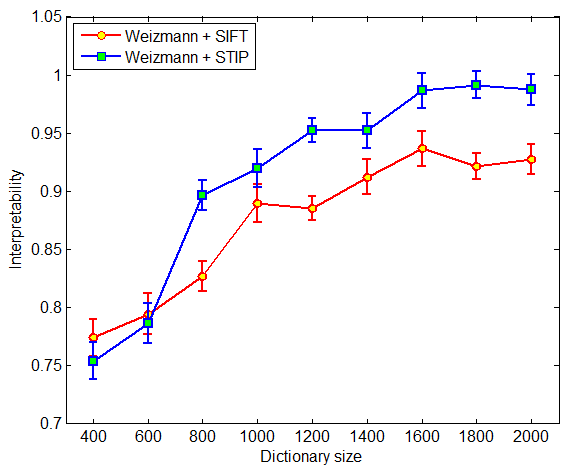}
    \end{minipage}}
  \subfigure[KTH dataset]{
    \label{fig:ii_kth}
    \begin{minipage}[b]{0.5\textwidth}
      \centering
        \includegraphics[width=1\textwidth]{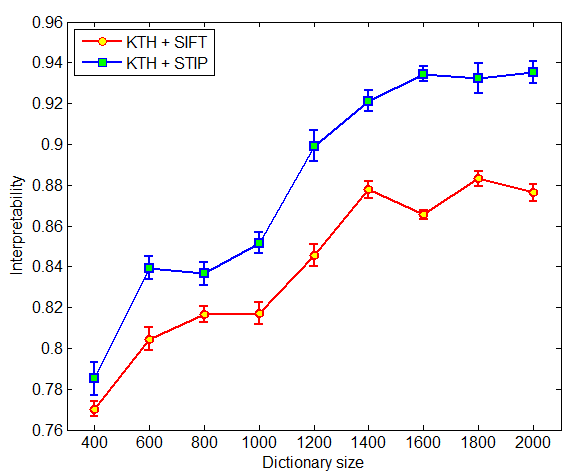}
    \end{minipage}}
      \subfigure[UTK3D dataset]{
    \label{fig:ii_utk3d}

    \begin{minipage}[b]{1\textwidth}
    \centering
        \includegraphics[width=0.5\textwidth]{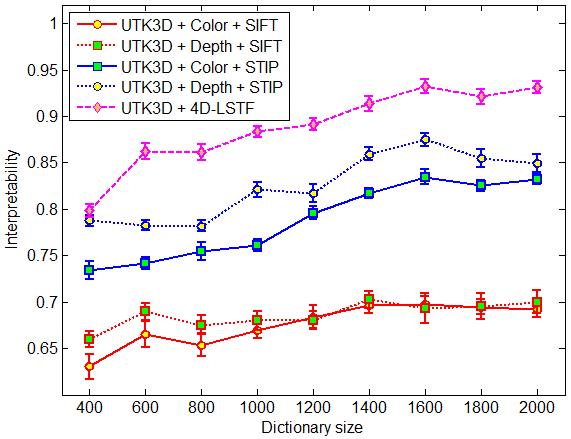}
    \end{minipage}}
  \caption{
Variations of model interpretability and its standard deviation versus dictionary size using different visual features over benchmark datasets.
}
\label{fig:ii}
\end{figure}

Experimental results of the interpretability and its standard deviation versus the dictionary size are illustrated in Fig. \ref{fig:ii}.
Our SRAC model obtains promising recognition performance in terms of interpretability:
0.989 is obtained using the STIP feature and a dictionary size 1800 on the Weizmann dataset, 0.952 using the STIP feature and a dictionary size 2000 on the KTH dataset,
and 0.936 using the 4D-LSTF feature and a dictionary size 1600 on the UTK3D dataset.
In general, STIP features perform better than SIFT features for color data,
and 4D-LSTF features perform the best for RGB-D visual data.
The dictionary size in the range [1500, 2000] can generally result in satisfactory human activity recognition performance.
The results are also very consistent,
as illustrated by the small error bars in Fig. \ref{fig:ii},
which demonstrates our interpretability indicator's consistency.

\begin{figure*}
\centering
\includegraphics[width= 1 \textwidth]{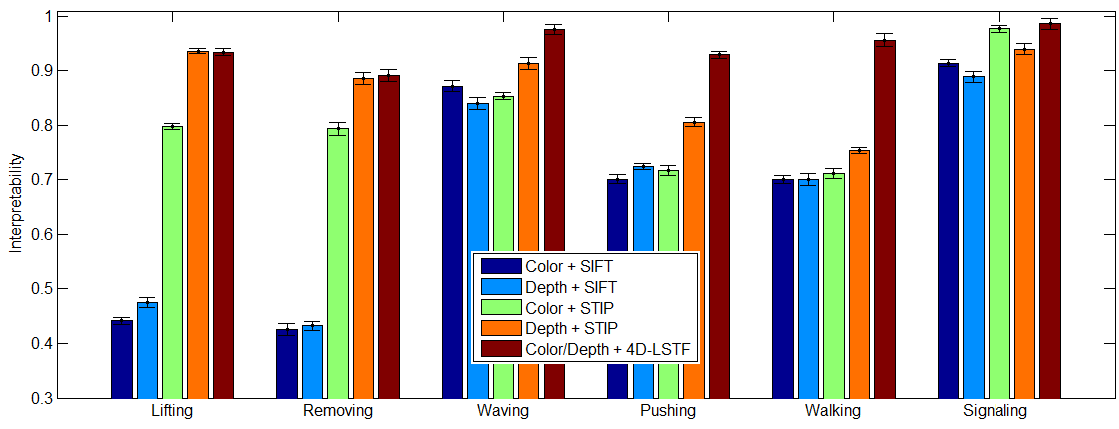}
\caption{
Model interpretability over the activities in the UTK3D dataset using different features and a dictionary size of 1600.
}\label{fig:ii_activity}
\end{figure*}

The model's interpretability is also evaluated over different activity categories
using the UTK3D dataset,
which includes more complex activities (i.e., sequential activities) and contains more information (i.e., depth).
It is observed that topic modeling's interpretability varies for different activities.
This performance variation is affected by three main factors:
the topic model's modeling ability, feature and BoW's representability,
and human activity complexity and similarity.
For example, since the LDA topic model and SIFT features are not capable of modeling time,
the reversal human activities including ``lifting a box'' and ``removing a box'' in the UTK3D dataset cannot be well distinguished, as illustrated in Fig. \ref{fig:ii_activity}.
Since sequential activities (e.g., ``removing a box'') are more complex
than repetitive activities (e.g., ``waving''),
they generally result in low interpretability.
Since ``pushing'' and ``walking'' are similar,
which share motions such as moving forward,
they can also reduce interpretability.
This observation provides general guidance for designing future recognition systems with the SRAC model.

\subsection{Knowledge Discovery} \label{sec:experiment-IG}

We evaluate the SRAC model's capability to discover new situations
using the generalizability indicator $I_G$,
when the training dataset is non-exhaustive (i.e., new human activities occur during testing).
A non-exhausted setup is created by dividing the used benchmark datasets as follows.
We place all data instances of one activity in the \emph{unknown testing set},
and randomly select 25\% of the instances from the remaining activities in the \emph{known testing set}.
The remaining instances are placed in the training set for learning,
based on fourfold cross-validation.
To evaluate the model's ability to discover each individual human activity,
given a dataset that contains $K$ activity categories,
the experiments are repeated $K$ times,
each using one category as the novel activity.
Visual features that achieve the best model interpretability over each dataset
are used in this set of experiments
i.e., STIP features for the Weizmann and KTH datasets
and 4D-LSTF features for the UTK3D dataset.



\begin{figure*}
  \subfigure[Weizmann dataset + STIP features]{
    \label{fig:pvwp_weizmann}
    \begin{minipage}[b]{0.5\textwidth}
      \centering
        \includegraphics[width=1.05\textwidth]{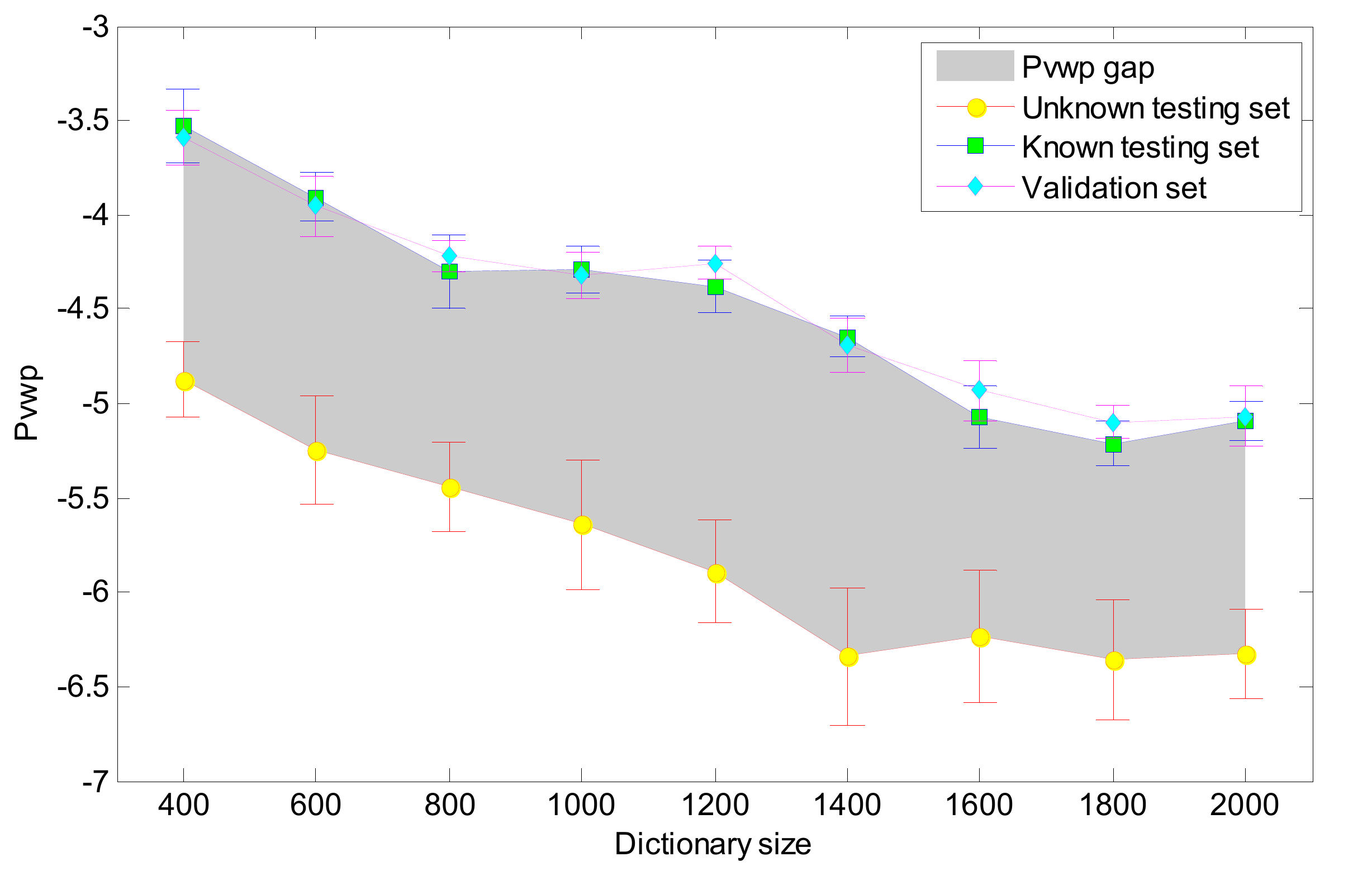}
    \end{minipage}}
  \subfigure[KTH dataset + STIP features]{
    \label{fig:pvwp_kth}
    \begin{minipage}[b]{0.5\textwidth}
      \centering
        \includegraphics[width=1.05\textwidth]{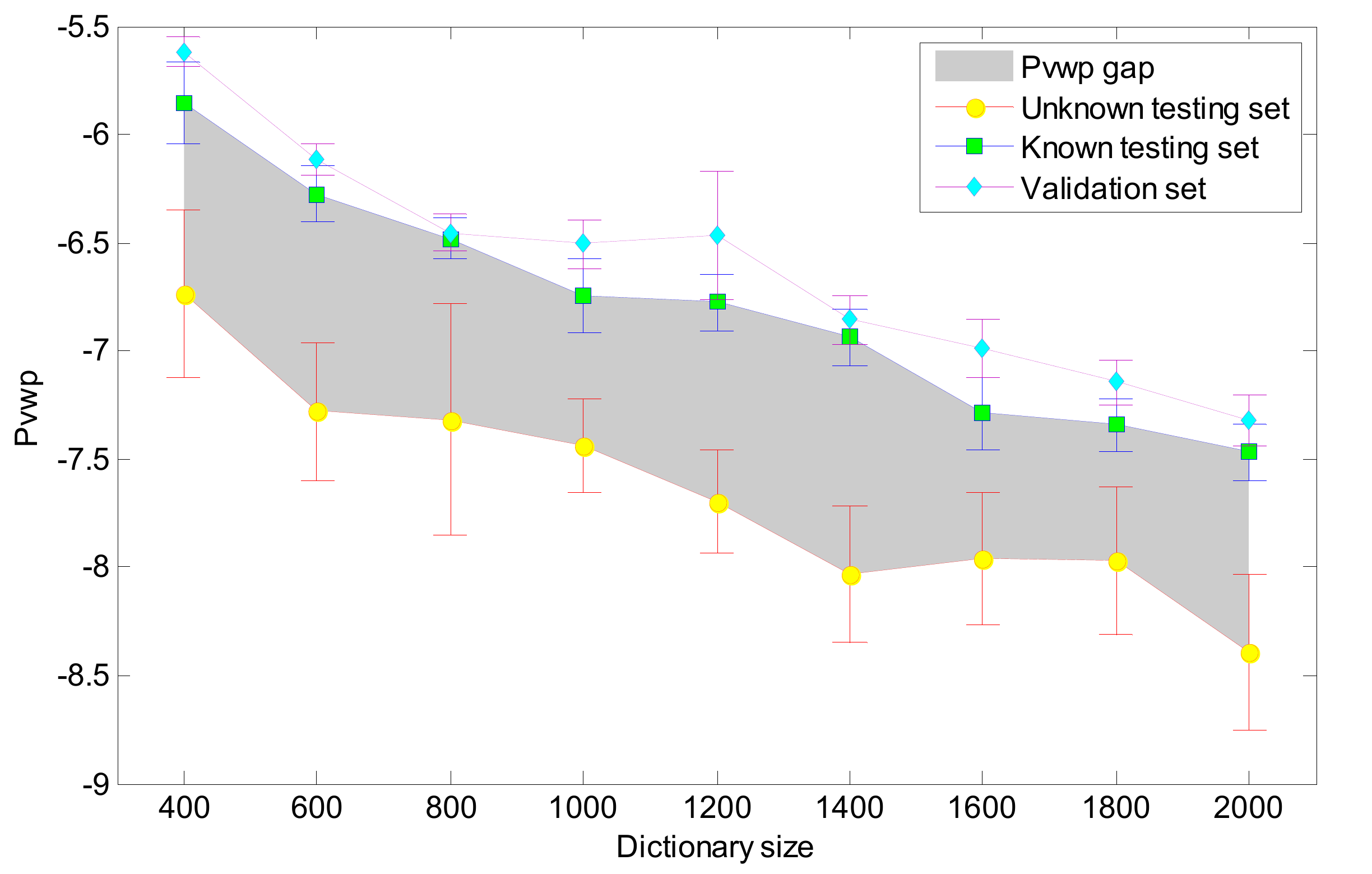}
    \end{minipage}}
      \subfigure[UTK3D dataset + 4D-LSTF features]{
    \label{fig:pvwp_utk3d}

    \begin{minipage}[b]{1\textwidth}
      \centering
        \includegraphics[width=0.5\textwidth]{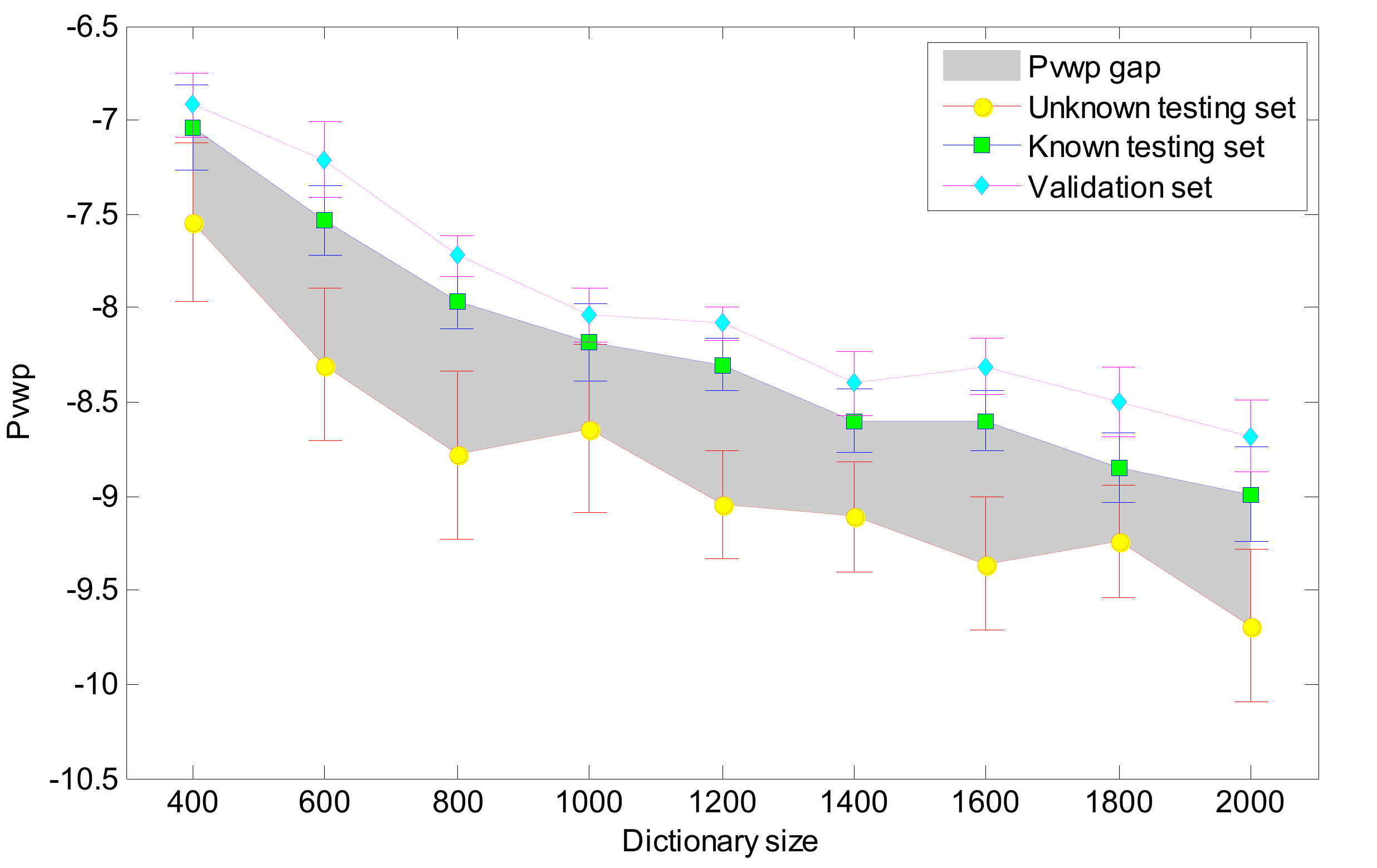}
    \end{minipage}}

  \caption{
Variations of topic modeling's $Pvwp$ versus dictionary size over validation set, known and unknown testing sets.
}
\label{fig:pvwp}
\end{figure*}

Variations of $Pvwp$ values versus the dictionary size over the validation set (in cross-validation),
known testing set, and unknown testing set
are shown in Fig. \ref{fig:pvwp}.
Several important phenomena are observed.
First,
there exists a large $Pvwp$ gap between the known and unknown testing sets,
as shown by the gray area in the figure, 
indicating that topic models
generalize differently over data instances
from known and unknown activities.
A better generalization result indicates a less novel instance,
which can be better represented by the training set.
Since data instances from the known testing and validation sets are well represented by the training set,
the $Pvwp$ gap between them is small.
As shown in Fig. \ref{fig:pvwp_weizmann},
it is possible that the known testing set's $Pvwp$ value
is greater than the $Pvwp$ value of the validation set,
if its data instances can be better represented by the training set.
Second, Fig. \ref{fig:pvwp}
 shows that the gap's width varies over different datasets:
the Weizmann dataset generally has the largest $Pvwp$ gap,
followed by the KTH dataset, and then the UTK3D dataset. 
The gap's width mainly depends on the observation's novelty,
in terms of the novel activity's similarity to the activities in the training dataset.
This similarity is encoded by the portion of overlapping features.
A more novel activity is generally represented by a set of more distinct visual features
with less overlapping with the features existing during training,
which generally results in a larger gap.
For example, activities in the Weizamann dataset share fewer motions
and thus contain a less number of overlapping features,
which leads to a larger gap.
Third, when the dictionary size increases,
the model's $Pvwp$ values decrease at a similar rate.
This is because in this case,
the probability of a specific codeword appearing in an instance decreases,
resulting in a decreasing $Pvwp$ value.


\begin{figure*}
  \subfigure[Weizmann dataset]{
    \label{fig:ig_weizmann}
    \begin{minipage}[b]{0.5\textwidth}
      \centering
        \includegraphics[width=1.05\textwidth]{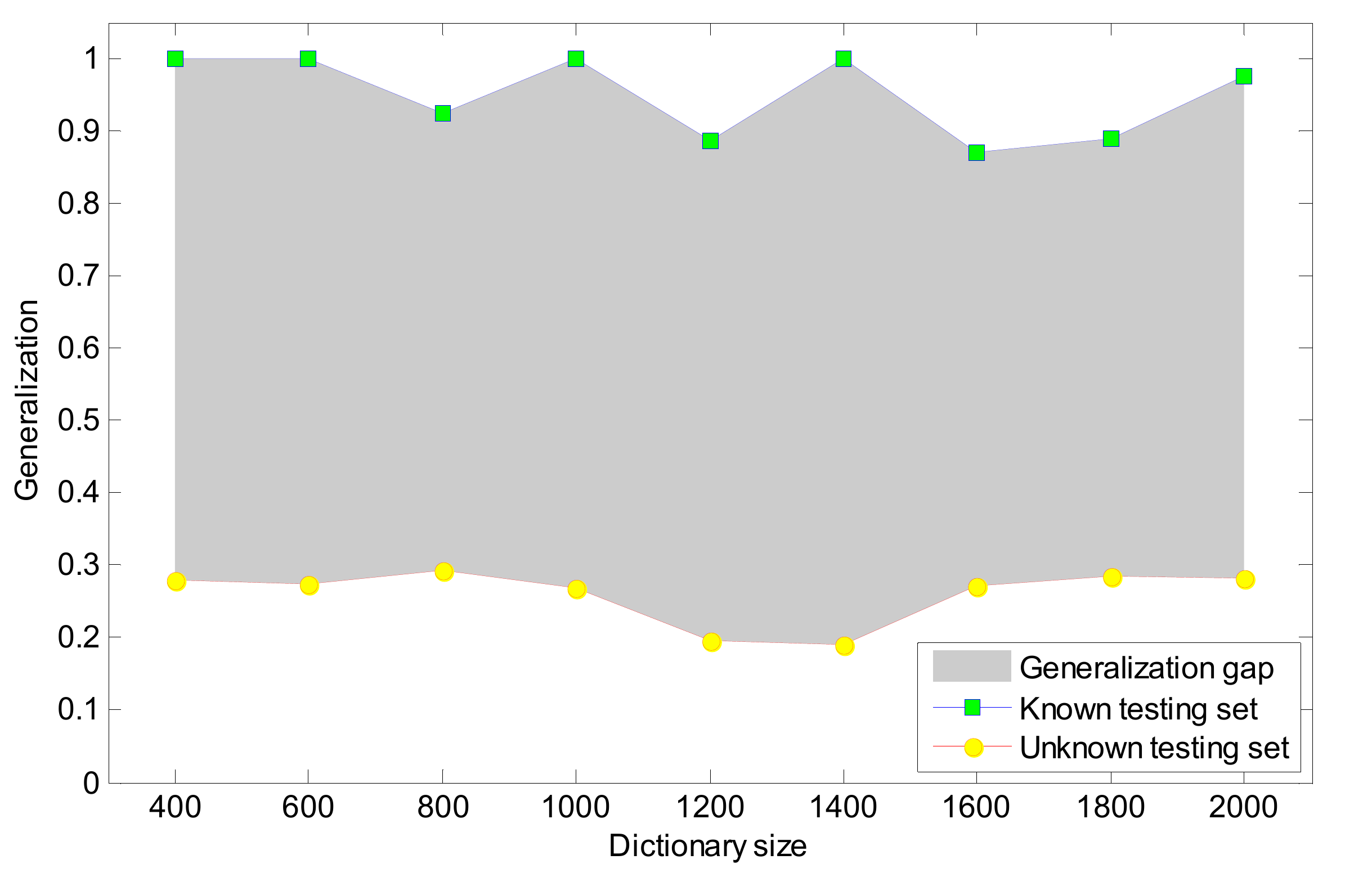}
    \end{minipage}}
  \subfigure[KTH dataset]{
    \label{fig:ig_kth}
    \begin{minipage}[b]{0.5\textwidth}
      \centering
        \includegraphics[width=1.05\textwidth]{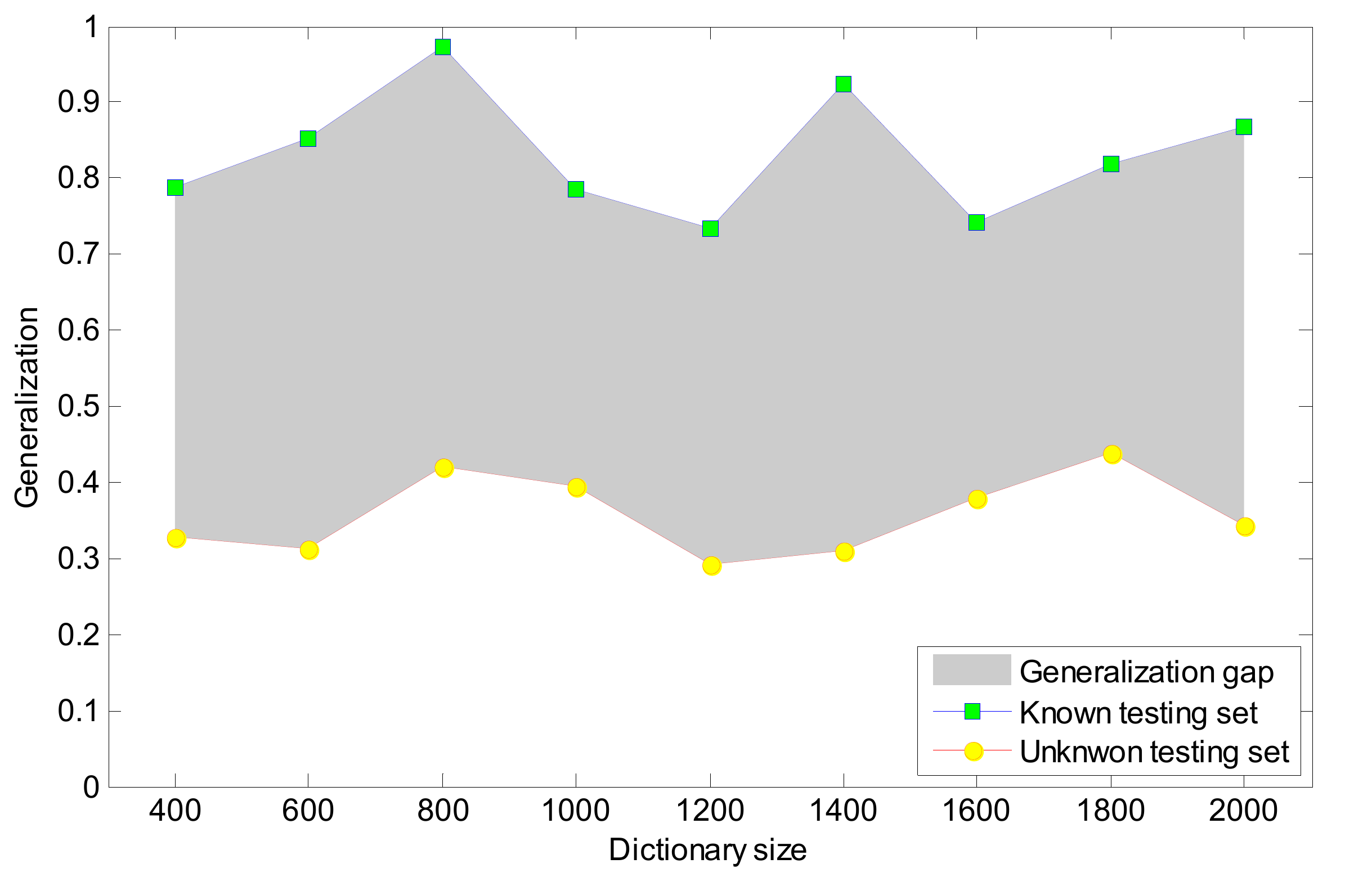}
    \end{minipage}}
      \subfigure[UTK3D dataset]{
    \label{fig:ig_utk3d}
    \begin{minipage}[b]{1\textwidth}
      \centering
        \includegraphics[width=0.505\textwidth]{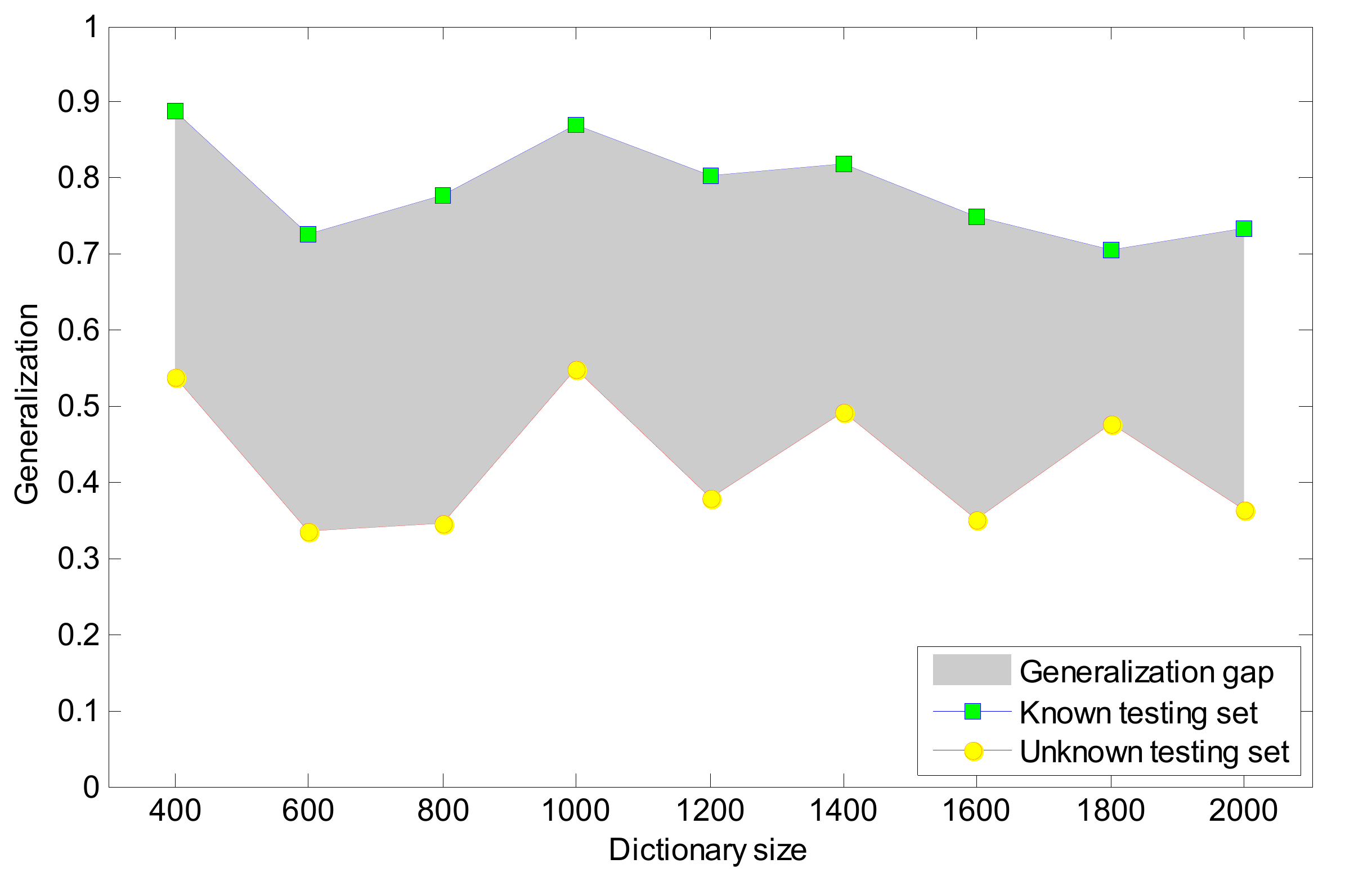}
    \end{minipage}}
  \caption{
Variations of our model's generalizability versus dictionary size over known and unknown testing sets for all datasets.
}\label{fig:ig}
\vspace{-6pt}
\end{figure*}

The generalizability indicator $I_G$'s characteristics
are also empirically validated on the known and unknown testing sets,
as illustrated in Fig. \ref{fig:ig}.
An important characteristic of $I_G$ is its invariance to dictionary size.
Because $Pvwp$ over testing and validation sets has similar decreasing rate,
the division operation in Eq. (\ref{eq:IG}) removes the variance to dictionary size.
In addition, a more novel activity generally leads to a smaller $I_G$ value.
For example, the Weizmann dataset has a smaller $I_G$ value over the unknown testing set,
because its activities are more novel in the sense that they share less overlapping motions.
In general, we observe $I_G$ is smaller than 0.5 for unknown activities 
and greater than 0.7 for activities that are included in training sets.
As indicated by the gray area in Fig. \ref{fig:ig},
similar to $Pvwp$,
there exists a large gap between the $I_G$ values over the unknown and known testing datasets.
The average $I_G$ gap across different dictionary sizes is $0.69$ for the Weizmann dataset,
$0.48$ for the KTH dataset, and $0.36$ for the UTK3D dataset.
This reasoning process, based on $I_G$, provides a co-robot with the critical self-reflection capability, and allows a robot to reason about when new situations occur as well as when the learned model becomes less applicable.

\begin{figure}
\centering
\includegraphics[width= 0.8 \textwidth]{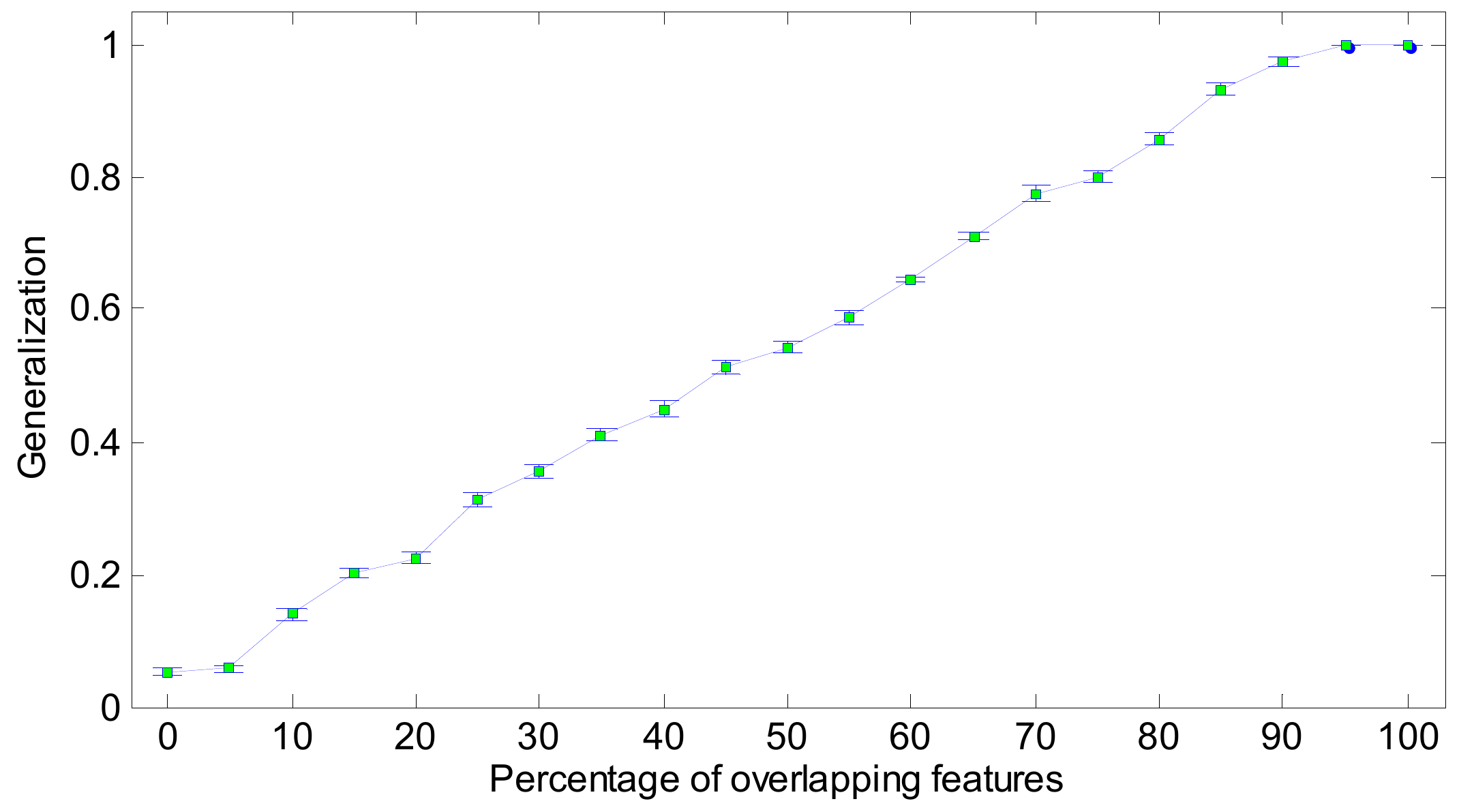}
\caption{
Variations of our model generalizability versus percentage of overlapping features in synthetic data.
}\label{fig:syn}
\end{figure}

We have pointed out that the indicator $I_G$ is heavily affected by the novelty of an activity
in terms of its proportion of overlapping features.
To validate this conclusion,
we generate a synthetic dataset
by manually controlling the proportion of overlapping visual words in the testing instances.
In order to make the characteristics of the synthetic dataset as close as possible to real-world datasets,
features used in the simulation are borrowed from the KTH dataset.
Instances of two activities (i.e., ``bending'' and ``waving2'')
are used to train a topic model, which is then applied as a classifier to perform recognition in this experiment.
This topic model is also applied to generate overlapping visual words for a testing instance.
Another topic model, whose parameters are learned using activities ``siding'' and ``jacking'',
is used to generate non-overlapping words in the testing instance.
A dictionary of size 1800 is adopted, which is created using the visual words of the KTH dataset.
The used four activities contain 906 unique visual words,
with each pair of activities sharing less than $1\%$ overlapping words.
We generate $50$ instances for each testing set,
with the number of words in each instance set to $112$,
which is the average number of visual words in real-world instances.
We present the results of five simulations in Fig. \ref{fig:syn},
which clearly shows that, in general, $I_G$'s value over testing instances increases linearly with
the percentage of features that overlap with the features of known activities in the training set.

\subsection{Relationship of $I_G$ and $I_I$}

Here,
we empirically analyze the relationship
between the interpretability and generalizability indicators.
We first validate the correlation of $I_I$ and $I_G$.
In addition,
we investigate additional relationships of $I_I$ and $I_G$, such as the probability that $I_I \leq I_G$.


While we are able to employ the exhaustive experimental setup from Section \ref{sec:experiment-II} to analyze $I_G$ and $I_I$'s relationship when testing instances are fully represented by the training set,
unfortunately,
we cannot use the non-exhaustive setup in Section \ref{sec:experiment-IG} to validate this relationship in cases where $I_G$ takes small values.
This is because ground truth cannot be assigned to instances belonging to novel activities to compute $I_I$,
since these activities only exist in the testing set
and are not presented to our model during the training phase.
Inspired by the method used to generate synthetic data in Section \ref{sec:experiment-IG},
we adopt a semi-exhaustive experimental setup by replacing certain portions of words in each testing instance with visual words from novel activities.
This experimental setup is used to validate the indicators' relationship when the training set cannot fully represent testing instances.

Each experiment is performed using $F$ folds, where $F$ is the number of activities in a dataset.
In each fold,
we take all instances of one activity out from the dataset, 
which is treated as a novel activity that is not presented to the topic model in the learning phase.
Then, we randomly select 75\% of the instances of the remaining activities as training set,
which is further divided into training and validation sets to perform four-fold cross-validation.
The rest of the instances are used as an ``initial'' testing set.
During the testing phase,
the novel activity's word distribution is used to generate new visual words to replace a proportion of the words in each instance in the initial testing set.
This testing is performed six times within each of the $F$ folds
using different replacement rates (i.e., $0.25$, $0.35$, $\dots$, $0.75$).
Testing results from all $F$ folds are used to investigate $I_I$ and $I_G$'s relationship.
In this experimental setup,
we use features that achieve the best interpretability over each dataset.
In addition, we set the dictionary size to 1600,
which achieves the best interpretability over all datasets in general.

This experimental setup is semi-exhaustive in the sense that,
although training data cannot fully represent testing instances due to the replaced features that are generated from unknown activities,
the remaining non-replaced features are presented to the model during the learning phase,
and the ground truth assigned to each testing instance remains the same, which is also known to the model.
It is noteworthy that we do not use very high or very low replacement rates.
A very low replacement rate makes the experimental setup equivalent to the exhaustive setup.
When using a very high replacement rate,
testing instances can be viewed as being drawn from the novel activity; in this case the ground truth associated with a testing instance would be meaningless or incorrect.

\begin{table}[tb]
\caption{
Relationship between $I_I$ and $I_G$
over exhaustive and semi-exhaustive datasets.
}
\label{tab:RelationIiIg}
\begin{center}
\begin{tabular}{|c|rc|cc|}
\hline
\multirow{2}{*}{Dataset + Features } & \multicolumn{2}{c|}{Exhaustive} & \multicolumn{2}{c|}{Semi-exhaustive} \\ \cline{2-3} \cline{4-5}
   & $\rho_{I,G}$ & $P_{I_I \!\leq\! I_G}$ & $\rho_{I,G}$  & $P_{I_I \!\leq\! I_G}$ \\
\hline\hline
Weizmann + STIP     & $-0.065$        & 0.456             & 0.664   & 0.912 \\
KTH + STIP          & $0.036$         & 0.324         & 0.685   & 0.853 \\
UTK3D + 4D-LSTF    & $0.097$         & 0.275         &  0.714    & 0.896 \\
\hline
\end{tabular}
\end{center}
\end{table}

We empirically analyze the correlation between $I_I$ and $I_G$,
using both exhaustive and semi-exhaustive datasets,
in order to determine whether better generalizability indicates better interpretability.
The Pearson correlation coefficient is used to measure the strength and direction of the linear relationship between these two indicators.
Given a dataset $\mathcal{W} = \{\boldsymbol{w}_1,\dots,\boldsymbol{w}_{\mathcal{|W|}}\}$ and its ground truth $\boldsymbol{g} = \{g_1,\dots,g_{\mathcal{|W|}}\}$,
this correlation is mathematically defined as follows:
\begin{eqnarray}
    \rho_{I,G} 
    = {E[(\boldsymbol{I}_I
     - \mu_{I_I})(\boldsymbol{I}_G -\mu_{I_G})] \over \sigma_{I_I}\sigma_{I_G}},
\end{eqnarray}
where $\boldsymbol{I}_G \!=\! \{I_G(\boldsymbol{w}_1), \dots, I_G(\boldsymbol{w}_{\mathcal{|W|}})\}$
and $\boldsymbol{I}_I \!=\!  \{I_I(\boldsymbol{w}_1, g_1),$
$\dots, I_I(\boldsymbol{w}_{\mathcal{|W|}}, g_{\mathcal{|W|}})\}$
are vectors of interpretability and generalizability indicators for all of the instances in the dataset,
$\mu$ is the mean and $\sigma$ is the standard deviation of the indicators in the vector.
Our experimental results are listed in Table \ref{tab:RelationIiIg}.
For the exhaustive dataset, topic models which perform better on generalizability are not necessarily better interpreted,
which is indicated by the weak linear correlation between the indicators. 
This is because, when testing on exhaustive datasets,
$I_G$ takes values closer to $1$. But the model's interpretability takes a wide range of values,
depending on the model's modeling capacity, feature representability and dataset complexity,
as explained in Section \ref{sec:experiment-II}.
For semi-exhaustive datasets,
$I_I$ and $I_G$ are moderately to strongly correlated,
which indicates that a poor generalizability usually leads to a poor interpretability.
Since $I_G$ reflects the novelty of an instance as discussed in Section \ref{sec:experiment-IG},
a low $I_G$'s value means the instance is badly represented by the training set.
Therefore, the trained model cannot obtain a good interpretability over the instance of an activity
that is not well represented during the training phase.

We also check an additional relationship, i.e., the probability that $I_I$ is smaller than or equal to $I_G$.
Given a labeled dataset $\mathcal{W} = \{\boldsymbol{w}_1,\dots,\boldsymbol{w}_M\}$
and its ground truth $\boldsymbol{g} = \{g_1,\dots,g_M\}$,
this probability is defined as follows:
\begin{eqnarray}
P_{I_I \leq I_G} = \frac{1}{M} \sum_{m = 1}^{M}
\mathds{1}(I_I(\boldsymbol{w}_m,g_m) \leq I_G(\boldsymbol{w}_m)).
\end{eqnarray}
The experimental results are presented in Table \ref{tab:RelationIiIg}.
One of the most important observations is that,
for a majority of testing instances (more than $85\%$) in the semi-exhaustive experiment,
$I_G$'s value is greater than $I_I$'s value.
This again shows that
a poor generalizability usually indicates a poor interpretability.
Using the exhaustive experimental setup,
it is more probable that $I_G$ takes smaller values than $I_I$.
This is because when the training set is exhaustive,
the topic model is well trained and can well recognize testing instances,
which leads to $I_I \!\rightarrow\! 1$ for most of testing instances.
On the other hand,
although $I_G$ also takes a large value in general,
it is usually slightly smaller than one, 
because features in testing instances usually do not completely overlap with features in training instances.

%
%


\subsection{Decision Making}

\begin{figure}
  \subfigure[Falling]{
    \label{fig:fall}
    \begin{minipage}[b]{0.32\textwidth}
      \centering
        \includegraphics[width=1\textwidth]{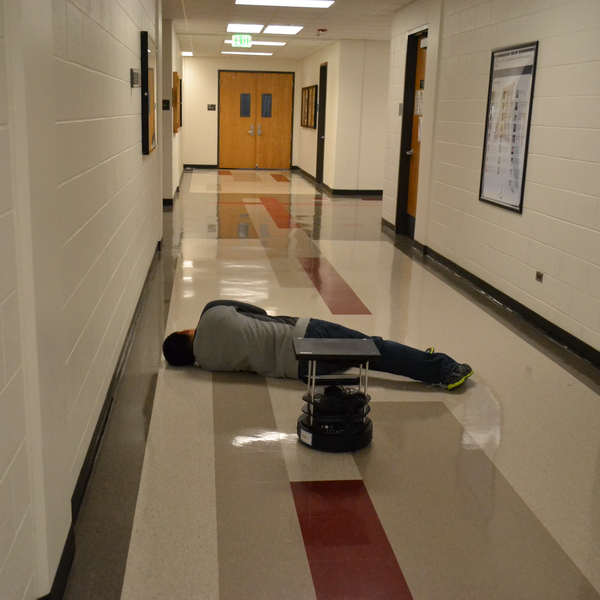}
    \end{minipage}}
  \subfigure[Turning]{
    \label{fig:turn}
    \begin{minipage}[b]{0.32\textwidth}
      \centering
        \includegraphics[width=1\textwidth]{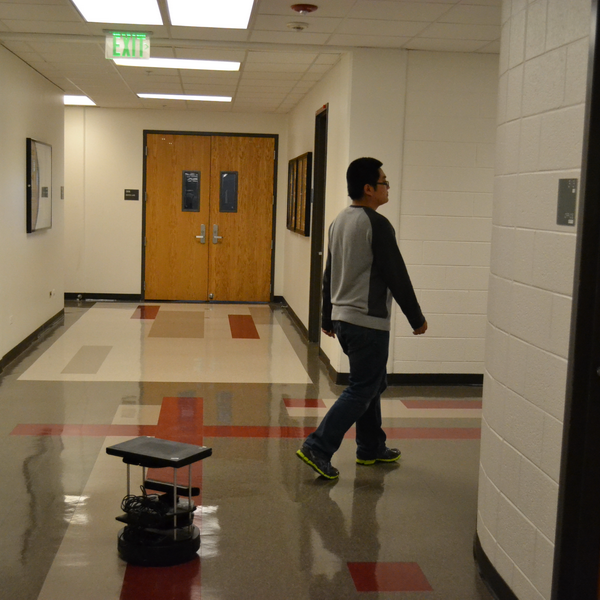}
    \end{minipage}}
      \subfigure[Walking]{
    \label{fig:walk}
    \begin{minipage}[b]{0.32\textwidth}
      \centering
        \includegraphics[width=1\textwidth]{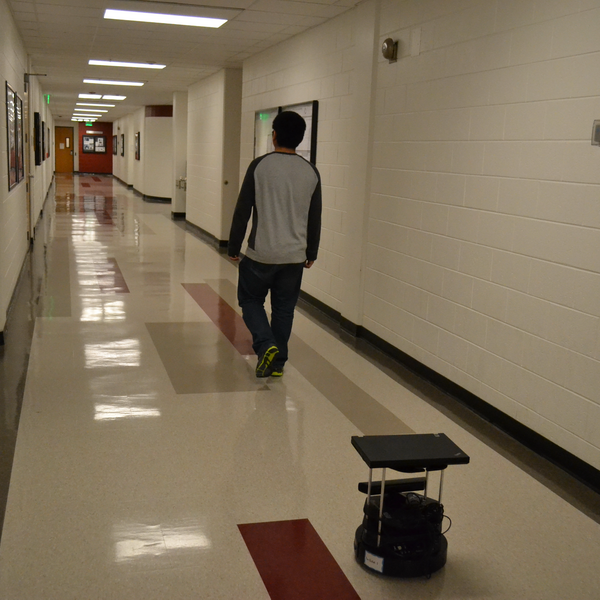}
    \end{minipage}}
  \caption{
Experiment setup for validating the SRAC model's decision making ability
in a human following task using a Turtlebot 2 robot.
}
\label{fig:DecisionEg}
\end{figure}

We assess our SRAC model's decision making capability
using a Turtlebot 2 robot in a human following task,
which is important in many human-robot teaming applications. 
In this task,
a robotic follower needs to
decide at what distance to follow the human teammate.
We are interested in three human behaviors:
``walking'' in a straight line, ``turning'', and ``falling'', shown in Fig. \ref{fig:DecisionEg}.
With perfect perception and reasoning,
i.e., a robot always perfectly interprets human activities,
we assume the ideal robot actions are to ``stay far from the human''
when he or she is walking in a straight line (to not interrupt the human),
``move close to the human'' when the subject is turning (to avoid losing the target),
and ``stop beside the human'' when he or she is falling (to provide assistance).

\begin{table}[t]
\centering
\caption{The risk matrix used in the robot following task.}
\label{tab:RiskMat}
\tabcolsep=0.35cm
\begin{tabular}{|c||c|c|c|}
\hline
Robot Actions      &	Falling  & Turning & Walking \\
\hline\hline
Stay besides humans  & 0 & 20 & 50 \\
\hline
Move close  & 90 & 0 & 20 \\
\hline
Stay far away & 95 & 80 & 0 \\
\hline
\end{tabular}
\end{table}

In order to qualitatively assess the performance,
we collect 20 color-depth instances from each human behaviors to train the SRAC model,
using a BoW representation based on 4D-LSTF features.
The risk matrix used in this task is presented in Table \ref{tab:RiskMat}.
We evaluate our model in two circumstances.
\texttt{Case$\,$1}: exhaustive training (i.e., no unseen human behaviors occur in testing).
In this case, the subjects only perform the three activities during testing with small variations in motion speed and style.
\texttt{Case$\,$2}: non-exhaustive training (i.e., novel movements occur during testing).
In this case, the subjects not only perform the activities with large variations,
but also add additional movements (such as jumping and squatting)
which are not observed in the training phase.
During testing,
each activity is performed 40 times.
The model performance is measured using failure rate,
i.e., the percentage with which the robot fails to stop besides to help the human
or loses the target.

Experimental results 
are presented in Table \ref{tab:ExpResults},
where the traditional methodology,
which selects the co-robot actions only based on the most probable human activity, is used as a baseline for comparison.
We observe that
the proposed SRAC model significantly decreases the failure rate
in both exhaustive and non-exhaustive setups.
When the training set is exhaustive and no new activities occur during testing (\texttt{Case$\,$1}),
the results demonstrate that
incorporating human activity distributions and robot action risks improves decision making performance.
When the training set is non-exhaustive and new activities occur during testing (\texttt{Case$\,$2}),
the SRAC model significantly outperforms the baseline model.
In this situation, if $I_G$ has a very small value,
according to Eq. \ref{eq:ActionSelection},
our model tends to select safer robot actions,
i.e., ``stay beside humans,'' since its average risk is the lowest,
which is similar to the human common practice ``playing it safe in uncertain times.''
The results show the importance of self-reflection for decision making especially under uncertainty.

\begin{table}[htbp]
\centering
\caption{Failure rate (\%) in exhaustive (\texttt{Case$\,$1}) and non-exhaustive (\texttt{Case$\,$2}) experimental settings.}
\label{tab:ExpResults}
\tabcolsep=0.35cm
\begin{tabular}{|c|c||c|c|}
\hline
Exp. settings & Models      &	Fail to assist  & Fail to follow \\
\hline\hline
Exhaustive  & Baseline & 10.5\% & 15\% \\
\cline{2-4}
(\texttt{Case$\,$1}) & \textbf{SRAC} &0.5\% & 5.5\% \\
\hline
Non-exhaustive  & Baseline & 45.5\% & 60\% \\
\cline{2-4}
(\texttt{Case$\,$2})  & \textbf{SRAC} & 24.5\% & 35.5\% \\
\hline
\end{tabular}
\end{table}

\section{Conclusion}\label{sec:Conclusion}

In this paper, we construct an artificial cognitive model that provides co-robots with both
accurate perception and new information discovery capabilities, which
enables safe, reliable robot decision making for HAR tasks in human-robot interaction applications.

The proposed SRAC model exploits topic modeling,
which is unsupervised and allows for the discovery of new knowledge without presence in training.
In addition, topic modeling is also able to treat activity estimation as a distribution and incorporate risks for each action response,
which is beneficial for the system's  ability to make decisions.
In order to provide the capability of accurate human activity interpretation,
we define a new interpretability indicator ($I_I$) and demonstrate
its ability to enable a robot to interpret category distribution in a similar fashion
to humans.
The indicator $I_I$ is applied to map detected clusters to known activity categories and select the best interpreted model.
In addition, to provide the ability of knowledge discovery,
we introduce a novel generalizability indicator ($I_G$). %
It measures how well an observation can be represented by the learned knowledge, which
allows for self-reflection that can enable the SRAC model to identify new scenarios.

We applied the proposed SRAC model in extensive experiments to demonstrate the effectiveness
using both synthetic and real-world datasets.
We show that our model performs extremely well in terms of interpretability;
that is, our model's recognition results closely and consistently match human common sense.
We demonstrate that, using $I_G$, our cognitive model is capable of discovering new knowledge,
i.e., observations from new activity categories that are not considered in the training phase can be automatically detected.
We also examine the relationship between $I_I$ and $I_G$ and show,
both analytically and experimentally,
that $I_G$ can also be used as an indicator for $I_I$.
The results reveal that scenarios with a low $I_G$ score for an observation will equate to a low $I_I$ score with high confidence,
i.e., a badly generalized model is likely to be inaccurate. 
We further demonstrate the advantages of using distributions over activity categories,
as well as the importance of the evaluation metrics
in order to create a system capable of safe, reliable decision making.


\appendix
\section{Proof of $I_I$'s Properties (Proposition 1)}\label{appendix:1}

If denominator in Definition 1 is 0, then limit is used.
Given the normalizing constants $a = 2$ and $b = 1$:

1. If $k = 1$,
$I_I(\boldsymbol{\theta}_s,k) =
\frac{1}{a} \left( 1 + b - \frac{\theta_2}{\theta_1}\right) = 1 - \frac{\theta_2}{2\theta_1}$.
Since $\boldsymbol{\theta}_s$ is decreasingly sorted, satisfying $\theta_1 \geq \theta_2 \geq 0$,
then $-\frac{\theta_2}{2\theta_1} \geq - 0.5$.
Thus, $I_I(\boldsymbol{\theta}_s,k) = 1 - \frac{\theta_2}{2\theta_1} \geq 0.5 $

2. If $k = K$,
$I_I(\boldsymbol{\theta}_s,k) =
\frac{1}{a} \left( \frac{\theta_K}{\theta_1} - 1 + b \right)
= \frac{\theta_K}{2\theta_1}$
Since $\boldsymbol{\theta}_s$ is decreasingly sorted, satisfying $\theta_1 \geq \theta_K \geq 0$,
then $\frac{\theta_K}{\theta_1} \leq 1$. Thus, $I_I(\boldsymbol{\theta}_s,k)
= \frac{\theta_K}{2\theta_1} \leq 0.5$.

3. First, we prove $I_I(\boldsymbol{\theta}_s,k) \geq 0$. Since $K \geq k > 0$ and $K \geq 2$,
the second multiplier $F_2 = \frac{K - k}{K - 1} + 1(k = K) > 0$.
Given $b = 1$,
the third multiplier satisfies
$F_3 = \frac{\theta_k}{\theta_1} - \frac{\theta_{k+{1}(k \neq K)}}{\theta_k} + b
= \frac{\theta_k^2 + \theta_1 (\theta_k - \theta_{k+{1}(k \neq K)})}{\theta_1\theta_k}$.
Since $\boldsymbol{\theta}_s$ is decreasingly sorted,
then $\theta_1 \geq \theta_k \geq \theta_{k + {1}(k = K)} \geq 0$.
Thus, $F_3 \geq 0$. Equality is obtained when $\theta_k = \theta_{k + {1}(k = K)} = 0$.
Since $a > 0$, $F_2 > 0$ and $F_3 \geq 0$, then
$I_I(\boldsymbol{\theta}_s,k) = \frac{1}{a} \cdot F_2 \cdot F_3 \geq 0$.
Now, we prove $I_I(\boldsymbol{\theta}_s,k) \leq 1$.
When $k = K$, by property 2, $I_I(\boldsymbol{\theta}_s,k) \leq 1$ directly holds.
If $K > k \geq 1$,
then $F_2 = \frac{K - k}{K - 1} \leq 1$.
Equality holds when $k = 1$.
Since $\boldsymbol{\theta}_s$ is decreasingly sorted,
satisfying $\theta_1 \geq \theta_k \geq \theta_{k+1} \geq 0$,
then $\frac{\theta_k}{\theta_1} \leq 1$ and $\frac{\theta_{k+1}}{\theta_k} \geq 0$.
Given $ b = 1$, we have $F_3 = \frac{\theta_k}{\theta_1} - \frac{\theta_{k+1}}{\theta_k} + b
\leq \frac{\theta_k}{\theta_1} + 1 \leq 2$.
Equality holds when $\theta_k = \theta_1$ and $\theta_{k+1} = 0$.
Thus, given $a = 2$, we obtain $I_I(\boldsymbol{\theta}_s,k) = \frac{1}{a}\cdot F_2 \cdot F_3 \leq 1$.
Thus, $\forall \boldsymbol{\theta}$, $I_I(\boldsymbol{\theta}_s,k) \in [0,1]$ holds.

4. Since $\forall k \in \{1,\cdots, K\}$,
$\boldsymbol{\theta}_s$, $\boldsymbol{\theta}'_s$ satisfy
$\theta_k = \theta_k'$ and $\theta_{k+ {l}(k = K)} = \theta_{k+ {l}(k = K)}'$,
we obtain
$I_I(\boldsymbol{\theta}_s,k) - I_I(\boldsymbol{\theta}'_s,k)
= \frac{1}{a} \left( \frac{K-k}{K-1} + {l}(k = K)\right)
\left(\frac{\theta_k}{\theta_1\theta_1'}(\theta_1' - \theta_1)\right)$.
Since $\frac{K-k}{K-1} + {l}(k = K) > 0$ and $\theta_1' \geq \theta_1$,
Then, $I_I(\boldsymbol{\theta}_s,k) - I_I(\boldsymbol{\theta}'_s,k) \leq 0$.
Equality holds if $\theta_1' = \theta_1$ or $\theta_k = 0$.
Thus,
$I_I(\boldsymbol{\theta}_s,k) \leq I_I(\boldsymbol{\theta}'_s,k)$.

5. Since $\forall k \in \{1,\cdots, K\}$,
$\boldsymbol{\theta}_s$, $\boldsymbol{\theta}'_s$ satisfy
$\theta_{1} = \theta_{1}'$ and $\theta_k = \theta_k'$,
we obtain
$I_I(\boldsymbol{\theta}_s,k) - I_I(\boldsymbol{\theta}'_s,k)
= \frac{1}{a} \left( \frac{K-k}{K-1} + {l}(k = K)\right)
\left(\frac{1}{\theta_k}(\theta_{k + {l}(k = K)}' -
\theta_{k + {l}(k = K)})\right)$.
Since $\frac{K-k}{K-1} + {l}(k = K) > 0$ and
$\theta_{k + {l}(k = K)} \geq
\theta_{k + {l}(k = K)}'$,
Then, $I_I(\boldsymbol{\theta}_s,k) - I_I(\boldsymbol{\theta}'_s,k) \leq 0$.
Equality holds if $\theta_{k + {l}(k = K)} =
\theta_{k + {l}(k = K)}'$.
Thus,
$I_I(\boldsymbol{\theta}_s,k) \leq I_I(\boldsymbol{\theta}'_s,k)$ holds.

6. Since $\forall k \in \{1,\cdots, K\}$,
$\boldsymbol{\theta}_s$, $\boldsymbol{\theta}'_s$ satisfy
$\theta_k = \theta_k'$ and $\theta_{k+ {l}(k = K)} = \theta_{k+ {l}(k = K)}'$,
we obtain
$I_I(\boldsymbol{\theta}_s,k) - I_I(\boldsymbol{\theta}'_s,k)
= \frac{1}{a} \left( \frac{K-k}{K-1} + {l}(k = K)\right)
\left(
\frac{1}{\theta_1} + \frac{\theta_{k + {l}(k = K)}}{\theta_k \theta_k'}
\right)
(\theta_k - \theta_k')
$.
Since $\frac{K-k}{K-1} + {l}(k = K) > 0$ and
$\frac{1}{\theta_1} + \frac{\theta_{k + {l}(k = K)}}{\theta_k \theta_k'} \geq 0$, and
$\theta_k > \theta_k'$,
Then, $I_I(\boldsymbol{\theta}_s,k) - I_I(\boldsymbol{\theta}'_s,k) \geq 0$.
Equality holds if $\theta_k = \theta_k'$.
Thus,
$I_I(\boldsymbol{\theta}_s,k) \geq I_I(\boldsymbol{\theta}'_s,k)$ holds.

7. $\forall k$, $k' \in \{1, \dots, K\}$ satisfying $k \leq k' < K$, and
$\forall \boldsymbol{\theta}_s$, $\boldsymbol{\theta}'_s$ satisfying $\theta_{k+1} = \theta_{k'+1}'$, $\theta_1 = \theta_1'$ and $\theta_k = \theta_{k'}'$,
we obtain
$I_I(\boldsymbol{\theta}_s,k) - I_I(\boldsymbol{\theta}'_s,k')
= \frac{1}{a(K-1)}
\left(\frac{\theta_k}{\theta_1} - \frac{\theta_{k + 1}}{\theta_k} + b \right)$
$(k' - k)$.
Since $K > 1$, $\frac{\theta_k}{\theta_1} - \frac{\theta_{k + 1}}{\theta_k} + b \geq 0$,
and $k' \geq k$,
$I_I(\boldsymbol{\theta}_s,k) - I_I(\boldsymbol{\theta}'_s,k') \geq 0$,
with equality holding when $\theta_k = \theta_{k + 1} = 0$ or $k = k'$.
Thus, $I_I(\boldsymbol{\theta}_s,k) \geq I_I(\boldsymbol{\theta}'_s,k')$ holds.


\section{Proof of the Relationship between $I_I$ and $I_A$ (Proposition 2)}\label{appendix:2}


Given an observation $\boldsymbol{w}$,
the accuracy metric $I_A$ indicates whether the recognition result $y(\boldsymbol{w})$ matches the ground truth $g$. Formally, $I_A$ is defined as follows:
\begin{eqnarray}\label{eq:p:IA}
I_A(y(\boldsymbol{w}), g) = {l}(y(\boldsymbol{w}) = g).
\end{eqnarray}

With this definition, we prove that $I_A$ is a special case of our $I_I$ indicator in Definition 1 when $\theta_1 = 1.0$, $\theta_2 =\! \dots \!=  \theta_K = 0$,
and $k = 1$ or $k = K$.

Given the normalizing constants $a = 2$ and $b = 1$,
when $\theta_1 = 1.0$, $\theta_2 =\! \dots \!=  \theta_K = 0$,
and $k = 1$ (i.e., the recognition result $y(\boldsymbol{w})$ matches the ground truth $g$), we obtain:
\begin{eqnarray*}\label{eq:p:II}
I_s(\boldsymbol{\theta}_s, 1) = \frac{1}{a}\left( \frac{K-1}{K-1} + 0 \right)
\left( \frac{\theta_1}{\theta_1} - \frac{\theta_2}{\theta_1} + b \right)
 = \frac{b+1}{a} = 1.
\end{eqnarray*}
When $k = K$ (i.e., $y(\boldsymbol{w})\neq g$), we obtain:
\begin{eqnarray*}
I_s(\boldsymbol{\theta}_s, K) \!=\! \frac{1}{a} \left(\frac{K\!-\!1}{K\!-\!1} \!+\! 1\right)
\left(\frac{\theta_K}{\theta_1} \!-\! \frac{\theta_K}{\theta_K} \!+\! b \right)
= \frac{2(b \!-\! 1)}{a} = 0.
\end{eqnarray*}
Combining both cases, we obtain:
\begin{eqnarray}
I_s(\boldsymbol{\theta}_s, k) &=&
\begin{cases}
1 & \textrm{if} \;\; k = 1 \;\; (\textrm{i.e.}, y(\boldsymbol{w}) = g)\\
0 & \textrm{if} \;\; k = K \;\; (\textrm{i.e.}, y(\boldsymbol{w}) \neq g)
\end{cases} \nonumber \\
&=& {l}(y(\boldsymbol{w}) = g).
\end{eqnarray}

We observe Eq. (\ref{eq:p:IA}) is equivalent to Eq. (\ref{eq:p:II}),
and thereby prove that $I_A$ is a special case of the $I_I$ indicator in the cases
when $\theta_1 = 1.0$, $\theta_2 =\! \dots \!=  \theta_K = 0$,
and $k = 1$ or $k = K$.

\bibliographystyle{elsarticle-num}
\bibliography{CogModel}

\begin{thebibliography}{10}
\expandafter\ifx\csname url\endcsname\relax
  \def\url#1{\texttt{#1}}\fi
\expandafter\ifx\csname urlprefix\endcsname\relax\def\urlprefix{URL }\fi
\expandafter\ifx\csname href\endcsname\relax
  \def\href#1#2{#2} \def\path#1{#1}\fi

\bibitem{Varela_99}
F.~J. Varela, J.~Dupuy, Understanding Origins, Kluwer Academic Publishers,
  1992.

\bibitem{Vernon_TEC07}
D.~Vernon, G.~Metta, G.~Sandini, A survey of artificial cognitive systems:
  Implications for the autonomous development of mental capabilities in
  computational agents, IEEE Transactions on Evolutionary Computation 11~(2)
  (2007) 151--180.

\bibitem{Anderson_AP96}
J.~R. Anderson, {ACT}: A simple theory of complex cognition, American
  Psychologist 51 (1996) 355--365.

\bibitem{trafton2013act}
G.~Trafton, L.~Hiatt, A.~Harrison, F.~Tamborello, S.~Khemlani, A.~Schultz,
  {ACT-R/E}: An embodied cognitive architecture for human-robot interaction,
  Journal of Human-Robot Interaction 2~(1) (2013) 30--55.

\bibitem{dancy2013act}
C.~L. Dancy, {ACT-R$\Phi$}: A cognitive architecture with physiology and
  affect, Biologically Inspired Cognitive Architectures 6 (2013) 40--45.

\bibitem{laird2012soar}
J.~Laird, The Soar cognitive architecture, MIT Press, 2012.

\bibitem{Isla_IJCAI01}
D.~Isla, R.~Burke, M.~Downie, B.~Blumberg, A layered brain architecture for
  synthetic creatures, in: International Joint Conferences on Artificial
  Intelligence, 2001.

\bibitem{Burghart_ICHR05}
C.~Burghart, R.~Mikut, R.~Stiefelhagen, T.~Asfour, H.~Holzapfel, P.~Steinhaus,
  R.~Dillmann, A cognitive architecture for a humanoid robot: a first approach,
  in: IEEE-RAS International Conference on Humanoid Robots, 2005.

\bibitem{Schmid_CogSysRes12}
U.~Schmid, M.~Ragni, C.~Gonzalez, J.~Funke, The challenge of complexity for
  cognitive systems, Cognitive Systems Research 12~(3-4) (2011) 211--218.

\bibitem{Le_CVPR11}
Q.~V. Le, W.~Y. Zou, S.~Y. Yeung, A.~Y. Ng, Learning hierarchical invariant
  spatio-temporal features for action recognition with independent subspace
  analysis, in: IEEE Conference on Computer Vision and Pattern Recognition,
  2011.

\bibitem{Alahi_CVPR12}
A.~Alahi, R.~Ortiz, P.~Vandergheynst, {FREAK}: Fast retina keypoint, in: IEEE
  Conference on Computer Vision and Pattern Recognition, 2012.

\bibitem{Zhang_IROS11}
H.~Zhang, L.~E. Parker, 4-dimensional local spatio-temporal features for human
  activity recognition., in: IEEE/RSJ International Conference on Intelligent
  Robots and Systems, 2011.

\bibitem{Girdhar_IJRR13}
Y.~Girdhar, P.~Giguere, G.~Dudek, Autonomous adaptive exploration using
  realtime online spatiotemporal topic modeling, International Journal of
  Robotics Research 33~(4) (2013) 645--657.

\bibitem{Niebles_IJCV08}
J.~C. Niebles, H.~Wang, L.~Fei-Fei, Unsupervised learning of human action
  categories using spatial-temporal words, International Journal of Computer
  Vision 79~(3) (2008) 299--318.

\bibitem{Aggarwal_CSUR11}
J.~Aggarwal, M.~Ryoo, Human activity analysis: A review, ACM Computing Surveys
  43~(3) (2011) 16:1--16:43.

\bibitem{borges2013video}
P.~V.~K. Borges, N.~Conci, A.~Cavallaro, Video-based human behavior
  understanding: a survey, IEEE Transactions on Circuits and Systems for Video
  Technology 23~(11) (2013) 1993--2008.

\bibitem{chen2014extraction}
X.~Chen, T.~Yang, Extraction method of gait feature based on human centroid
  trajectory, in: Computer Engineering and Networking, 2014, pp. 515--523.

\bibitem{ge2012vision}
W.~Ge, R.~T. Collins, R.~B. Ruback, Vision-based analysis of small groups in
  pedestrian crowds, IEEE Transactions on Pattern Analysis and Machine
  Intelligence 34~(5) (2012) 1003--1016.

\bibitem{Singh_2008}
M.~Singh, A.~Basu, M.~Mandal, Human activity recognition based on silhouette
  directionality, IEEE Transactions on Circuits and Systems for Video
  Technology 18~(9) (2008) 1280--1292.

\bibitem{junejo2014silhouette}
I.~N. Junejo, K.~N. Junejo, Z.~Al~Aghbari, Silhouette-based human action
  recognition using {SAX-Shapes}, The Visual Computer 30~(3) (2014) 259--269.

\bibitem{zhang2015bio}
H.~Zhang, L.~E. Parker, Bio-inspired predictive orientation decomposition of
  skeleton trajectories for real-time human activity prediction, in: IEEE
  International Conference on Robotics and Automation, 2015.

\bibitem{Lowe_IJCV04}
D.~G. Lowe, Distinctive image features from scale-invariant keypoints,
  International Journal of Computer Vision 60~(2) (2004) 91--110.

\bibitem{Sun_CVPR09}
X.~Sun, M.~Y. Chen, A.~Hauptmann, Action recognition via local descriptors and
  holistic features, in: IEEE Conference on Computer Vision and Pattern
  Recognition, 2009.

\bibitem{behera2014real}
A.~Behera, A.~G. Cohn, D.~C. Hogg, Real-time activity recognition by discerning
  qualitative relationships between randomly chosen visual features, in:
  British Machine Vision Conference, 2014.

\bibitem{Schuldt_ICPR04}
C.~Schuldt, I.~Laptev, B.~Caputo, Recognizing human actions: A local {SVM}
  approach, in: International Conference on Pattern Recognition, 2004.

\bibitem{Dollar_ICCV05}
P.~Doll\'ar, V.~Rabaud, G.~Cottrell, S.~Belongie, Behavior recognition via
  sparse spatio-temporal features, in: IEEE International Workshop on Visual
  Surveillance and Performance Evaluation of Tracking and Surveillance, 2005.

\bibitem{zhao2013relevance}
F.~Zhao, Y.~Huang, L.~Wang, T.~Tan, Relevance topic model for unstructured
  social group activity recognition, in: Advances in Neural Information
  Processing Systems, 2013.

\bibitem{Wang_PAMI09}
Y.~Wang, G.~Mori, Human action recognition by semilatent topic models, IEEE
  Transactions on Pattern Analysis and Machine Intelligence 31~(10) (2009)
  1762--1774.

\bibitem{Huynh_DAP08}
T.~Huynh, M.~Fritz, B.~Schiele, Discovery of activity patterns using topic
  models, in: International Conference on Ubiquitous Computing, 2008.

\bibitem{Farrahi_TIST11}
K.~Farrahi, D.~Gatica-Perez, Discovering routines from large-scale human
  locations using probabilistic topic models, ACM Transactions on Intelligent
  Systems and Technology 2~(1) (2011) 3:1--3:27.

\bibitem{freedman2014temporal}
R.~G. Freedman, H.-T. Jung, S.~Zilberstein, Temporal and object relations in
  plan and activity recognition for robots using topic models, in: AAAI Fall
  Symposium Series, 2014.

\bibitem{Wallach_ICML2009}
H.~Wallach, I.~Murray, R.~Salakhutdinov, D.~Mimno, Evaluation methods for topic
  models, in: International Conference on Machine Learning, 2009.

\bibitem{Wei_ICRDIR06}
X.~Wei, W.~B. Croft, {LDA}-based document models for ad-hoc retrieval, in:
  Interational Conference on Research and Development in Information Retrieval,
  2006.

\bibitem{chang_nips09}
J.~Chang, J.~Boyd-Graber, S.~Gerrish, C.~Wang, D.~Blei, Reading tea leaves: How
  humans interpret topic models, in: Neural Information Processing Systems,
  2009.

\bibitem{Newman_HLT10}
D.~Newman, J.~H. Lau, K.~Grieser, T.~Baldwin, Automatic evaluation of topic
  coherence, in: Human Language Technologies: The Annual Conference of the
  North American Chapter of the Association for Computational Linguistics,
  2010.

\bibitem{Blei_ComACM12}
D.~M. Blei, Probabilistic topic models, Communications of the ACM 55~(4) (2012)
  77--84.

\bibitem{Gray_HCI97}
W.~D. Gray, R.~M. Young, S.~S. Kirschenbaum, Introduction to this special issue
  on cognitive architectures and human-computer interaction, Human-Computer
  Interaction 12~(4) (1997) 301--309.

\bibitem{Duric_IEEE02}
Z.~Duric, W.~Gray, R.~Heishman, F.~Li, A.~Rosenfeld, M.~Schoelles, C.~Schunn,
  H.~Wechsler, Integrating perceptual and cognitive modeling for adaptive and
  intelligent human-computer interaction, Proceedings of the IEEE 90~(7) (2002)
  1272--1289.

\bibitem{lallee2014efaa}
S.~Lall{\'e}e, V.~Vouloutsi, S.~Wierenga, U.~Pattacini, P.~Verschure, {EFAA}: a
  companion emerges from integrating a layered cognitive architecture, in:
  ACM/IEEE International Conference on Human-robot Interaction, 2014.

\bibitem{baxter2013cognitive}
P.~E. Baxter, J.~de~Greeff, T.~Belpaeme, Cognitive architecture for
  human--robot interaction: towards behavioural alignment, Biologically
  Inspired Cognitive Architectures 6 (2013) 30--39.

\bibitem{Yang_TST13}
B.~Yang, L.~Zhou, Z.~Deng, {C-HMAX}: Artificial cognitive model inspired by the
  color vision mechanism of the human brain, Tsinghua Science and Technology
  18~(1) (2013) 51--56.

\bibitem{halbrugge2013act}
M.~Halbr{\"u}gge, {ACT-CV}: Bridging the gap between cognitive models and the
  outer world, Grundlagen und anwendungen der mensch-maschine-interaktion
  (2013) 205--210.

\bibitem{Nagel_AIM04}
H.-H. Nagel, Steps toward a cognitive vision system, AI Magine 25~(2) (2004)
  31--50.

\bibitem{lawitzky2013interactive}
A.~Lawitzky, D.~Althoff, C.~F. Passenberg, G.~Tanzmeister, D.~Wollherr,
  M.~Buss, Interactive scene prediction for automotive applications, in:
  Intelligent Vehicles Symposium, 2013, pp. 1028--1033.

\bibitem{Crowley_CVS06}
J.~L. Crowley, Things that see: Context-aware multi-modal interaction., in:
  Cognitive Vision Systems, 2006, pp. 183--198.

\bibitem{brdiczka2009learning}
O.~Brdiczka, J.~L. Crowley, P.~Reignier, Learning situation models in a smart
  home, IEEE Transactions on Systems, Man, and Cybernetics, Part B: Cybernetics
  39~(1) (2009) 56--63.

\bibitem{Xu_Cog11}
F.~Xu, T.~L. Griffiths, Probabilistic models of cognitive development: Towards
  a rational constructivist approach to the study of learning and development,
  Cognition 120 (2011) 299--301.

\bibitem{bencomo2013dynamic}
N.~Bencomo, A.~Belaggoun, V.~Issarny, Dynamic decision networks for
  decision-making in self-adaptive systems: A case study, in: International
  Symposium on Software Engineering for Adaptive and Self-Managing Systems,
  2013.

\bibitem{kafai2012dynamic}
M.~Kafai, B.~Bhanu, Dynamic {Bayesian} networks for vehicle classification in
  video, IEEE Transactions on Industrial Informatics 8~(1) (2012) 100--109.

\bibitem{Chater_TCS06}
N.~Chater, J.~B. Tenenbaum, A.~Yuille, Probabilistic models of cognition: where
  next?, Trends in Cognitive Sciences 10~(7) (2006) 292--293.

\bibitem{Blei_JMLR03}
D.~M. Blei, A.~Y. Ng, M.~I. Jordan, Latent dirichlet allocation, Journal of
  Machine Learning Research 3 (2003) 993--1022.

\bibitem{Griffiths_NAS04}
T.~L. Griffiths, M.~Steyvers, Finding scientific topics, in: National Academy
  of Sciences, 2004.

\bibitem{Porteous_KDD08}
I.~Porteous, D.~Newman, A.~Ihler, A.~Asuncion, P.~Smyth, M.~Welling, Fast
  collapsed {Gibbs} sampling for latent dirichlet allocation, in: ACM SIGKDD
  International Conference on Knowledge Discovery and Data Mining, 2008.

\bibitem{Musat_IJCAI11}
C.~C. Musat, J.~Velcin, S.~Trausan-Matu, M.-A. Rizoiu, Improving topic
  evaluation using conceptual knowledge, in: International Joint Conference on
  Artificial Intelligence, 2011.

\bibitem{Blei_NIPS05}
D.~Blei, J.~Lafferty, Correlated topic models, in: Neural Information
  Processing Systems, 2006.

\bibitem{Dundar_ICML12}
M.~Dundar, F.~Akova, A.~Qi, B.~Rajwa, Bayesian nonexhaustive learning for
  online discovery and modeling of emerging classes, in: International
  Conference on Machine Learning, 2012.

\bibitem{Gorelick_PAMI07}
L.~Gorelick, M.~Blank, E.~Shechtman, M.~Irani, R.~Basri, Actions as space-time
  shapes, IEEE Transactions on Pattern Analysis and Machine Intelligence
  29~(12) (2007) 2247--2253.

\bibitem{Laptev_IJCV05}
I.~Laptev, On space-time interest points, International Journal of Computer
  Vision 64~(2-3) (2005) 107--123.

\end{thebibliography}

\end{document}